\documentclass[10pt,twocolumn,letterpaper]{article}

\usepackage[accsupp]{axessibility}
\usepackage[pagenumbers]{cvpr} 

\definecolor{cvprblue}{rgb}{0.21,0.49,0.74}
\usepackage[pagebackref,breaklinks,colorlinks,allcolors=cvprblue]{hyperref}
\usepackage{tikz}
\usepackage{amsmath}
\usepackage{amssymb}
\usepackage{dsfont}

\def\paperID{5617} 
\def\confName{CVPR}
\def\confYear{2025}

\title{TKG-DM: Training-free Chroma Key Content Generation Diffusion Model}

\author{
Ryugo Morita$^{1, 2}$, Stanislav Frolov$^{2}$, Brian Bernhard Moser$^{2}$, Takahiro Shirakawa,\\ Ko Watanabe$^{2}$, Andreas Dengel$^{2}$, Jinjia Zhou$^{1}$ \\
$^{1}$Faculty of Science and Engineering, Hosei University, Tokyo, Japan \\
$^{2}$RPTU Kaiserslautern-Landau \& DFKI GmbH, Kaiserslautern, Germany \\
}

\begin{document}
\maketitle

\begin{abstract}
Diffusion models have enabled the generation of high-quality images with a strong focus on realism and textual fidelity. Yet, large-scale text-to-image models, such as Stable Diffusion, struggle to generate images where foreground objects are placed over a chroma key background, limiting their ability to separate foreground and background elements without fine-tuning.
To address this limitation, we present a novel Training-Free Chroma Key Content Generation Diffusion Model (TKG-DM), which optimizes the initial random noise to produce images with foreground objects on a specifiable color background. 
Our proposed method is the first to explore the manipulation of the color aspects in initial noise for controlled background generation, enabling precise separation of foreground and background without fine-tuning.
Extensive experiments demonstrate that our training-free method outperforms existing methods in both qualitative and quantitative evaluations, matching or surpassing fine-tuned models.
Finally, we successfully extend it to other tasks (e.g., consistency models and text-to-video), highlighting its transformative potential across various generative applications where independent control of foreground and background is crucial.
\end{abstract}
    
\begin{figure}[t]
    \centering
    \includegraphics[width=0.46\textwidth]{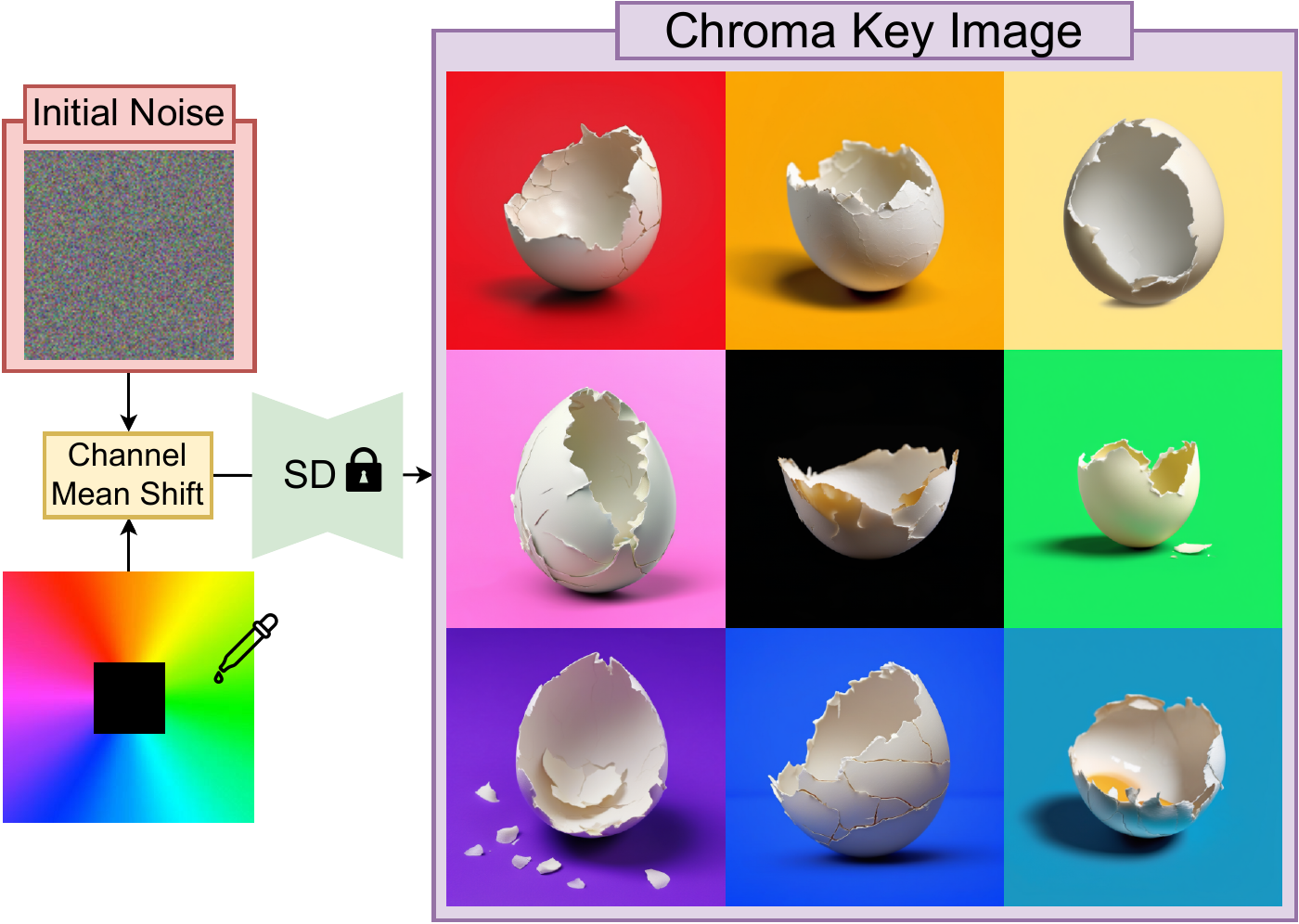} 
    \caption{TKG-DM provides a training-free generation of foreground objects on a chroma key background, enabling independent control of foreground and background elements. Additionally, our method extends to conditional text-to-image and video generation tasks, making it suitable for creating foreground assets for various applications.}
    \label{fig:fig1}
\end{figure}

\section{Introduction}
\label{sec:intro}
Diffusion models have disrupted the landscape of image generation by producing high-quality images with high realism and textual fidelity \cite{sohl2015deep, dhariwal2021diffusion, saharia2022photorealistic, chefer2023attend, moser2024diffusion}. 
Research has expanded into text-to-image generation conditioned on inputs like edges \cite{zhang2023adding, voynov2023sketch}, layouts \cite{zheng2023layoutdiffusion, yang2023reco}, and Virtual Try-ON (VTON) \cite{xu2024ootdiffusion, choi2024improving}. 
However, real-world applications such as advertising, design, and game development often require precise control over foreground and background elements. The ability to generate images with transparent or chroma key backgrounds is essential for seamlessly integrating foreground objects into various scenes \cite{zhang2024transparent}.
Traditional models like Stable Diffusion \cite{rombach2022high} and DeepFloyd \cite{saharia2022photorealistic} struggle to generate images where foreground and background elements are separated without fine-tuning \cite{burgert2024magick}. Specifically, these models cannot produce foreground objects over specifiable backgrounds or generate independent foreground and background layers, limiting their applicability in workflows requiring such precise control.

Recent works like MAGICK \cite{burgert2024magick} and LayerDiffuse \cite{zhang2024transparent} have significantly addressed this limitation. 
MAGICK introduces a prompt engineering-based approach for chroma key image generation. 
However, due to limited chroma key accuracy, it relies on post-processing and manual efforts to improve quality, including the release of foreground alpha image datasets. 
LayerDiffuse proposes a fine-tuning method for layer-wise generation using a dataset with 1 million images. 
Unfortunately, this dataset is not publicly available due to licensing restrictions, necessitating further resource-intensive dataset curation and fine-tuning.

To overcome these challenges, we present the Training-Free Chroma Key Content Generation Diffusion Model (TKG-DM), which optimizes the initial noise in the diffusion process to produce images with foreground objects on specifiable color backgrounds without fine-tuning. 
Our method is the first to explore manipulating the color aspects of initial noise for controlled background generation, enabling precise separation of foreground and background elements.
TKG-DM offers high flexibility and provides precise control over the background color, layout, size, and number of foreground objects. 
Our approach expands possibilities in generative content by maintaining the independence of the foreground and chroma key background. 
It seamlessly extends to applications like conditional text-to-image generation, consistency models, and text-to-video generation.

Our extensive experiments demonstrate that TKG-DM improves FID and mask-FID scores by 33.7\% and 35.9\%, respectively. 
Thus, our training-free model rivals fine-tuned models, offering an efficient and versatile solution for various visual content creation tasks that require precise foreground and background control.
Our contributions are summarized as follows:

\begin{itemize}
    \item We introduce TKG-DM, a training-free diffusion model that eliminates the need for fine-tuning and datasets for controllable foreground and background generation.
    \item TKG-DM provides precise control over background color, as well as the size, position and number of foreground elements, enabling flexible monochromatic chroma key generation.
    \item Our method is highly versatile, seamlessly extending to conditional text-to-image generation, consistency models, and text-to-video applications, thus enabling the creation of diverse chroma key content.
    \item TKG-DM outperforms existing models, delivering results that match or surpass fine-tuned alternatives while maintaining lower computational costs.
\end{itemize}


\section{Related Works}
\label{sec:related_works}

\subsection{Diffusion Models}
\label{sec:diffusion_model}
Diffusion models \cite{ho2020denoising, nichol2021improved} generate realistic, high-quality images by iteratively refining Gaussian noise to approximate a target distribution \cite{dhariwal2021diffusion, song2020denoising}. 
Building upon this, Latent Diffusion Models (LDMs) \cite{rombach2022high} improve efficiency by exploiting the low-dimensional latent spaces while preserving image quality. 
In particular, text-to-image models \cite{nichol2021glide, ramesh2021zero, saharia2022photorealistic, rombach2022high} align text and image features through cross-attention to achieve diverse, high-quality outputs, with classifier-free guidance \cite{ho2022classifier} and attention control \cite{chefer2023attend, cao2023masactrl} further enhancing text fidelity. 
Beyond image generation, diffusion models find use in inpainting \cite{yang2023paint, lugmayr2022repaint}, image editing \cite{hertz2022prompt, brooks2023instructpix2pix}, and layout generation \cite{inoue2023layoutdm, chai2023layoutdm}. 
Their versatility extends across multiple modalities, including audio generation \cite{huang2023make, kong2020diffwave}, text-to-video \cite{guo2023animatediff, wu2023tune}, and 3D object generation \cite{qian2023magic123, voleti2024sv3d}.

\subsection{Initial Noise}
\label{sec:init_noise}
Diffusion models typically start with Gaussian noise and iteratively refine it to generate high-quality images that align closely with a given target distribution. 
In general, the choice and optimization of this initial noise are crucial, as they significantly impact image quality and alignment \cite{xu2024good, samuel2024generating, samuel2024norm}.
More specifically, noise optimization methods, such as human preference models or attention-based scores, can enhance fidelity and text alignment without additional training \cite{eyring2024reno, guo2024initno}. 
Moreover, optimal noise seeds and targeted noise regions significantly impact image quality and object placement \cite{ban2024crystal, xu2024good}. 
In contrast, noise manipulation enables precise layout control, supporting layout-aware image generation and editing \cite{shirakawa2024noisecollage, mao2023semantic, mao2023guided}. 
Building on these insights, our work introduces a novel method for shifting initial noise to control chromatic information.
As a result, our method enables accurate chroma key backgrounds while preserving high-quality foreground content.

\subsection{Foreground and Background Separation in Image Generation}
\label{sec:layerd_generation}
Foreground-background separation is crucial in advertising, design, and game development, allowing creators to manipulate individual image elements independently for tasks such as background replacement, foreground adjustments, and complex compositions \cite{zhang2023text2layer, huang2024layerdiff, chen2024anyscene, benny2020onegan, morita2023batinet, morita2023interactive, dalva2024layerfusion, quattrini2025alfie, zou2025zero}.

In response to this challenge, MAGICK \cite{burgert2024magick} employs a prompt engineering-based model to solve this task. However, as noted in this work: ``Unfortunately, all the publicly available methods we tested were incapable of consistently creating keyable images.'' 
To overcome this limitation, MAGICK combines multiple models and applies manual effort to release a foreground alpha image dataset. 
Similarly, LayerDiffuse \cite{zhang2024transparent} introduces a fine-tuning approach using a 1 million image dataset, which cannot be publicly shared due to licensing restrictions. 
Although these methods improve foreground-background separation, they require substantial data collection and computational resources, making them less efficient for real-time applications. 
In contrast, our training-free approach provides precise control over foreground and background elements without the need for extensive datasets and fine-tuning.
\begin{figure*}[t]
    \centering
    \includegraphics[width=\textwidth]{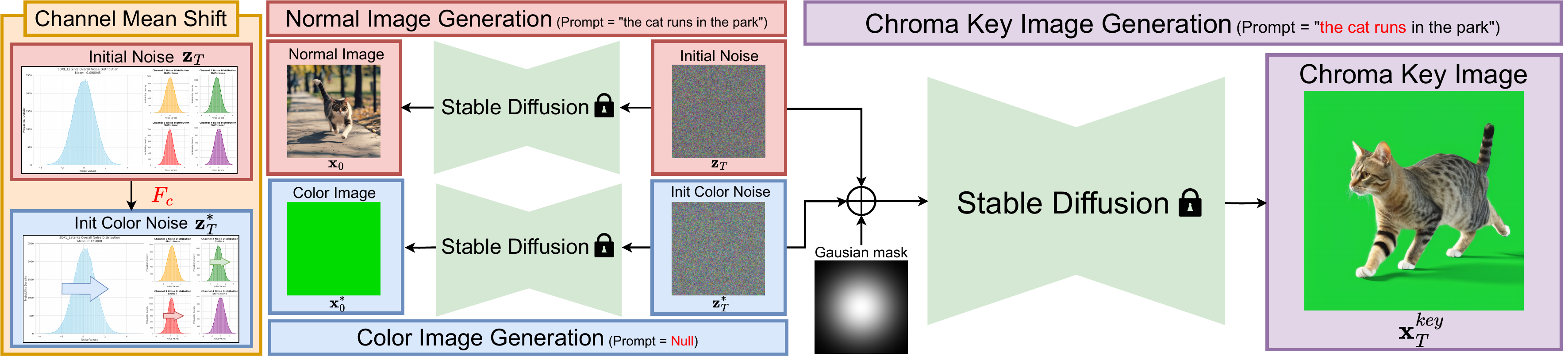} 
    \caption{Starting with random noise $\mathbf{z}_T \sim \mathcal{N}(\mathbf{0}, \mathbf{I})$, init color noise $\mathbf{z}_T^*$ is generated by applying channel mean shift $F_c$. This produces a single-colored image $\mathbf{x}_0^*$ without a prompt. By combining normal noise with init color noise via a Gaussian mask, TKG-DM generates a chroma key image $\mathbf{x}_0^{key}$ with the specified foreground (e.g., ``the cat runs in the park'') over a uniform background, effectively separating the foreground from the monochromatic background.}
    \label{fig:model}
\end{figure*}


\section{Methodology} 
\label{sec:foreground-gen}
As shown in Fig. \ref{fig:model}, TKG-DM extends Stable Diffusion \cite{rombach2022high} by manipulating the initial Gaussian noise $\mathbf{z}_T \in \mathbb{R}^{h \times w \times 4}$ through \textit{channel mean shift} $F_c$. This transformation produces the \textit{init color noise} $\mathbf{z}_T^* = F_c(z_T)$, which guides the vanilla Stable Diffusion to generate a uniform color image $\mathbf{x}_0^*$ without any text prompt.
To generate an image where the foreground aligns with the input text prompt $p$ and the background has a specified color (e.g., green for chroma keying), we combine the initial noise $\mathbf{z}_T$ and the init color noise $\mathbf{z}_T^*$ using a Gaussian mask. This combined noise $\mathbf{z}_T^{key}$ is input into the vanilla Stable Diffusion, generating the chroma key image $\mathbf{x}_0^{key}$.

\subsection{Channel Mean Shift} 
\label{sec:channe_mean_shift}
Inspired by the relationship between Stable Diffusion's latent space and generated image color~\cite{vass2023sdxl}, we introduce a novel initial noise optimization technique, \textit{channel mean shift}. It adjusts the mean of each channel in $z_T$ while keeping its standard deviation constant, enabling control over the generated image's color. To determine the optimal shift $\Delta_c$ for each channel $c \in \{1, 2, 3, 4\}$ to achieve a target positive ratio in that channel.
The values of channel $c$ in the noise tensor are denoted as $\mathbf{z}_T^{(c)}$. The initial positive ratio for channel $c$ is defined as:
\begin{equation}
    \text{InitialRatio}_c = \frac{\sum_{i,j} \mathds{1}(\mathbf{z}_T^{(c)}(i,j) > 0)}{\text{TotalPixels}_c},
\end{equation}
\noindent
where $\mathds{1}(\cdot)$ is the indicator function, and $\text{TotalPixels}_c$ is the total number of elements in $\mathbf{z}_T^{(c)}$.
Given a target shift $\text{TargetShift}_c$, the target positive ratio is:
\begin{equation}
\text{TargetRatio}_c = \text{InitialRatio}_c + \text{TargetShift}_c.
\end{equation}
To achieve $\text{TargetRatio}_c$, we iteratively adjust the mean shift $\Delta_c$ for each channel $\mathbf{z}_T^{(c)}$. We initialize the shift with $\Delta_c^{\text{init}} = 0$ and incrementally adjust $\Delta_c$ until the positive ratio meets or exceeds $\text{TargetRatio}_c$. Once the target ratio is reached, we record the final shift as $\Delta_c^{\text{final}} = \Delta_c$.
The noise tensor obtained through this method is called \textit{init color noise} $\mathbf{z}_T^* = F_c(\mathbf{z}_T) = \mathbf{z}_T + \Delta_c^{\text{final}}$.

\subsection{Init Noise Selection} 
\label{sec:init_selector}
To generate the foreground object on the chroma key background, we apply an init noise selection strategy that selectively combines the initial noise $\mathbf{z}_T$ and the init color noise $\mathbf{z}_T^*$ using a 2D Gaussian mask $\mathbf{A}(i, j)$. This mask creates a gradual transition by preserving the original noise in the foreground region and applying the color-shifted noise to the background region.
Formally, the masked initial noise is computed as:
\begin{equation} 
{\mathbf{z}_T^{key}}(i, j) = \mathbf{A}(i, j) \cdot \mathbf{z}_T(i, j) + (1 - \mathbf{A}(i, j)) \cdot {\mathbf{z}_T}^*(i, j),
\end{equation} 
\noindent
where $\mathbf{A}(i, j) = e^{-\frac{(i - \mu_i)^2 + (j - \mu_j)^2}{2\sigma^2}}$ controls the blend between the original noise and init color noise and $\mu_i$ and $\mu_j$ specify the center and $\sigma$ the spread. 

Adjusting the Gaussian parameters $\mu_i$, $\mu_j$, and $\sigma$ controls the position and size of the foreground: shifting $\mu_i$ and $\mu_j$ adjusts its position while increasing $\sigma$ enlarges the foreground element. 
Interestingly, TKG-DM also generates multiple foreground elements by applying multiple Gaussian masks, each independently configured to produce varied positions and sizes for flexible compositions.
\begin{figure*}[t]
    \centering
    \begin{subfigure}[b]{0.6\textwidth}
        \centering
        \includegraphics[width=\textwidth]{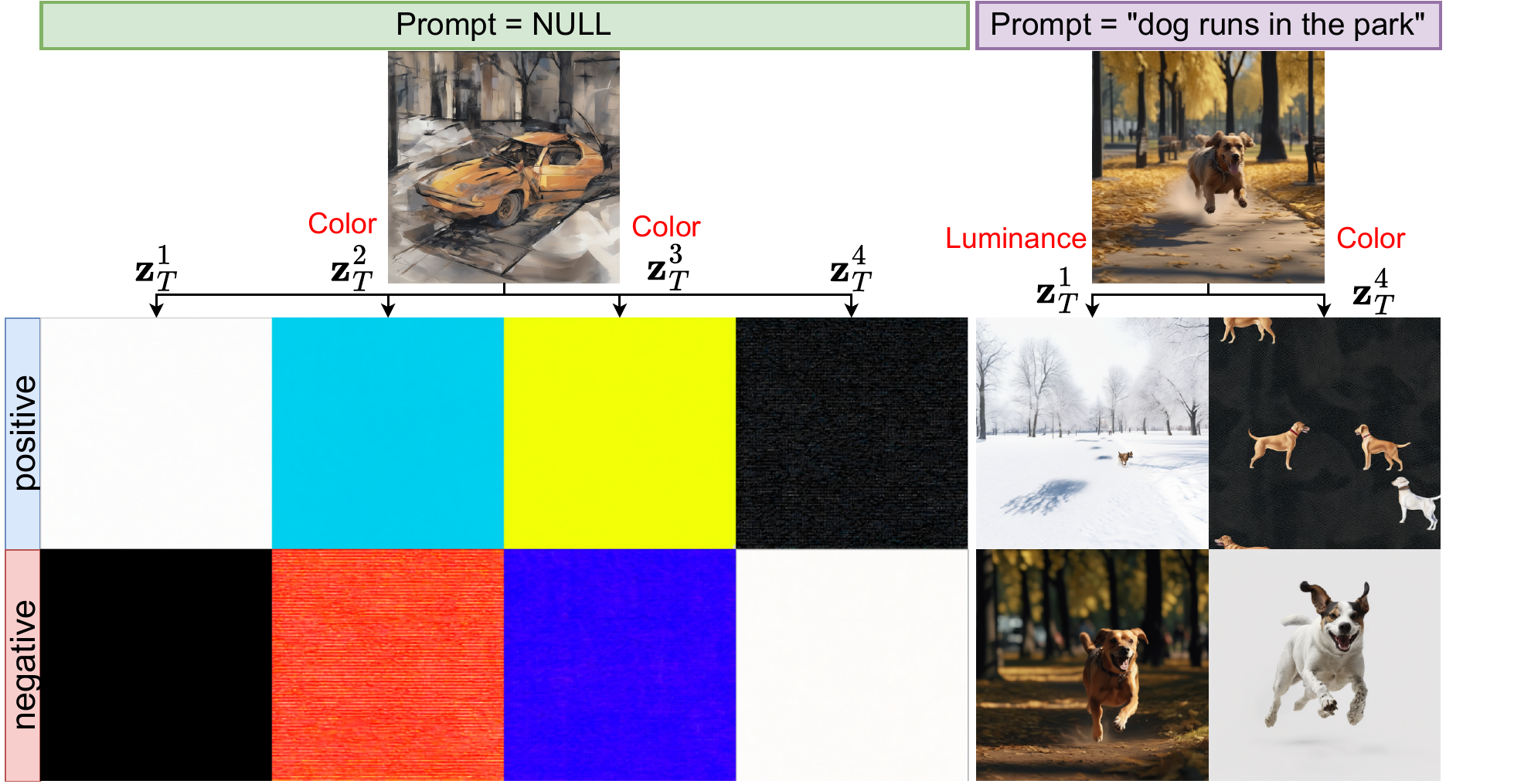} 
        \caption{Influence of Individual Noise Channel}
        \label{fig:influence}
    \end{subfigure}%
    \hspace{0.01\textwidth}
    \begin{tikzpicture}
    \draw[dotted] (2,0) -- (2,6.45);
    \end{tikzpicture}
    \hspace{0.01\textwidth}
    \begin{subfigure}[b]{0.356\textwidth}
        \centering
        \includegraphics[width=\textwidth]{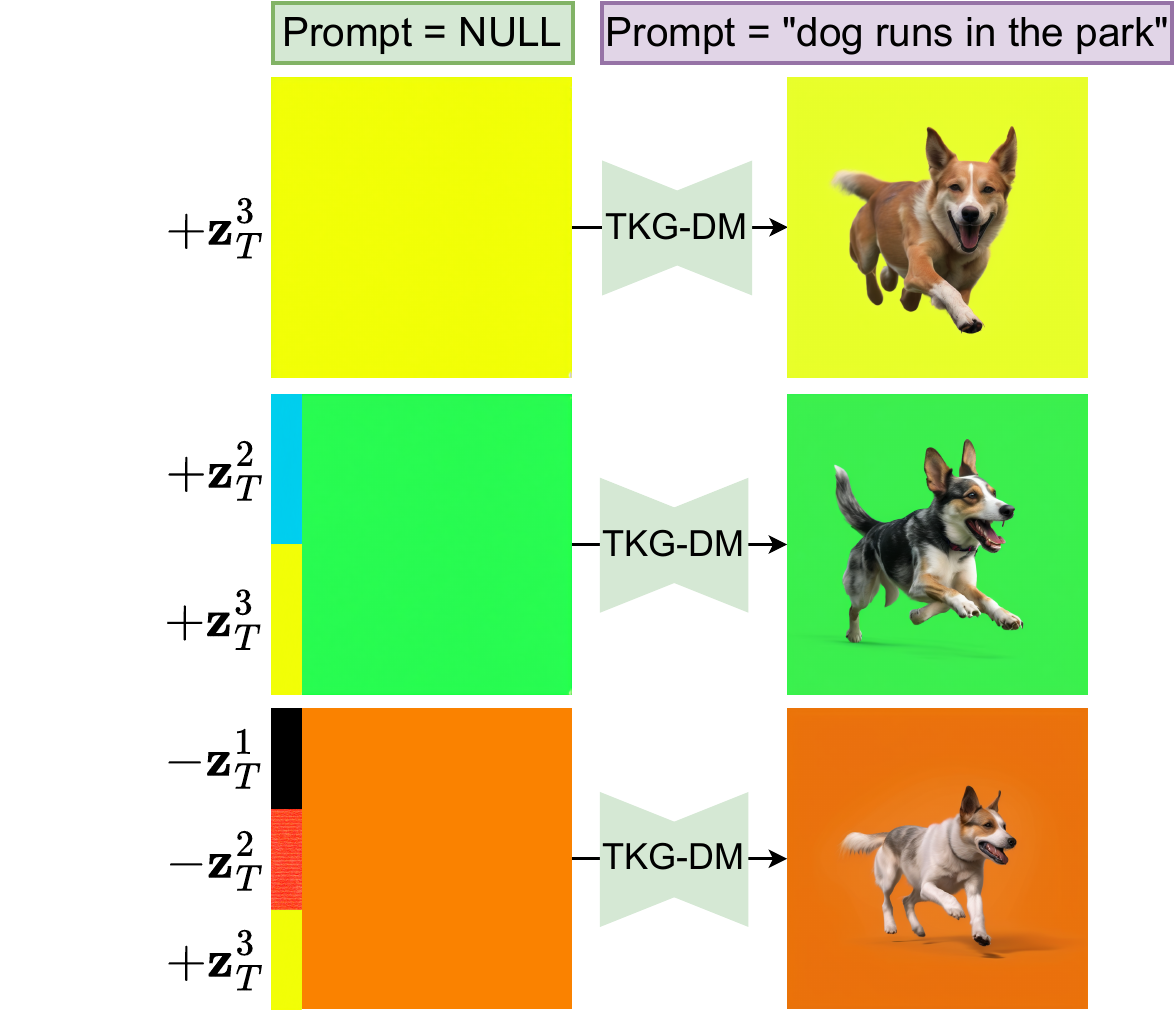} 
        \caption{Additive and Subtractive Color Effects}
        \label{fig:add}
    \end{subfigure}
    \caption{Fig.~\ref{fig:influence} illustrates the relationship between positive/negative channel mean shift with initial noise and color variations in the generated images. Different channel shifts across noise channels result in changes in the hue of the generated image. Fig.~\ref{fig:add} shows how simultaneous shifts across multiple channels facilitate additive and subtractive color mixing, providing intuitive and flexible color control.}
    \label{fig:analysis}
\end{figure*}

\section{Mechanism of TKG-DM}
\label{sec:mechanism}
In this section, we explain how TKG-DM specifies the background color and generates the foreground and background separately to generate chroma key content.

\subsection{Color Specify via Channel Mean Shift}
\label{sec:analysis}
Inspired by previous research~\cite{vass2023sdxl}, we control the chroma key background color by applying channel mean shift to specific channels of the initial noise $\mathbf{z}_T \in \mathbb{R}^{h \times w \times 4}$. Specifically, we adjust the mean of each channel $\mathbf{z}_T^{(c)}$, where $c \in \{1,2,3,4\}$, to influence the color composition of the generated images. In this experiment, we set $\text{TargetShift}_c = \pm7\%$, adjusting the positive ratio of each channel by adding or subtracting $7\%$ relative to $\text{InitialRatio}_c$.

As shown in Fig.~\ref{fig:influence}, positive shifts in $\mathbf{z}_T^{(2)}$ and $\mathbf{z}_T^{(3)}$ intensify cyan and yellow tones, respectively, while negative shifts emphasize red and blue-purple hues. Shifts in $\mathbf{z}_T^{(1)}$ and $\mathbf{z}_T^{(4)}$ primarily affect luminance and shades of white and black, respectively. These effects are confirmed by comparing generated images conditioned on specific text prompts.

When we apply simultaneous shifts to multiple channels, we observe intuitive color mixing that aligns with additive and subtractive color theories. For example, positive shifts in both $\mathbf{z}_T^{(2)}$ and $\mathbf{z}_T^{(3)}$ enhance green tones, while combining red and yellow shifts with reduced luminance yields orange hues (see Fig.~\ref{fig:add}). This channel-based manipulation enables flexible and natural color control in the image generation process. 

\subsection{Content Separation via Attention Mechanism}
\begin{figure}[t]
    \centering
    \includegraphics[width=0.47\textwidth]{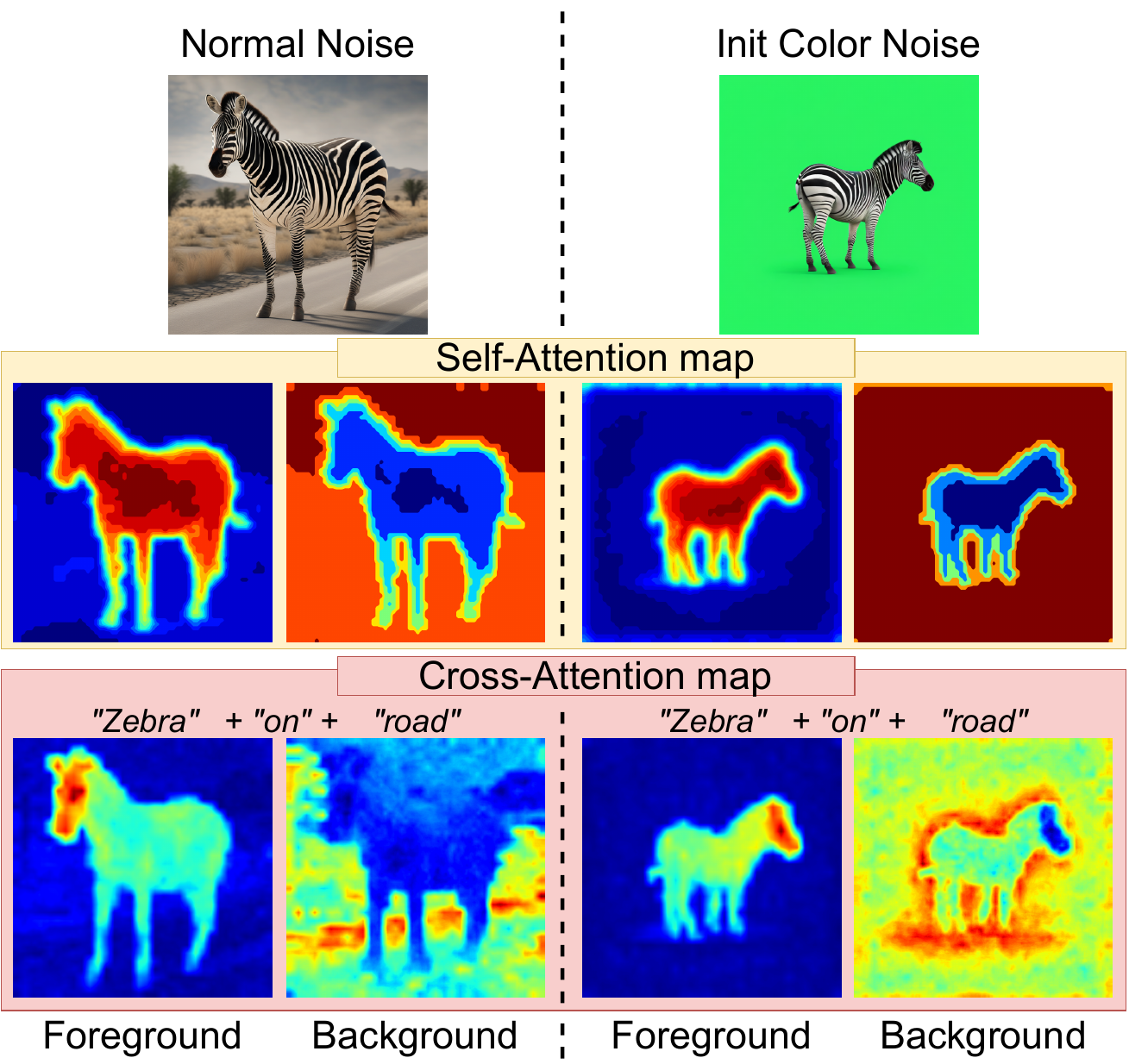}
    \caption{\textbf{Visualization of self- and cross-attention maps.} The self-attention maps, segmented using a foreground segmentation model \cite{zheng2024bilateral}, show the regions attended to during generation. The cross-attention maps illustrate how the model attends to relevant regions based on the foreground and background prompts. This analysis shows how self-attention and cross-attention interact to influence foreground and background content generation.}
    \label{fig:atten}
\end{figure}

As illustrated in Fig.~\ref{fig:atten}, self-attention and cross-attention mechanisms play distinct roles in generating the foreground and background elements, which is crucial for achieving effective foreground-background separation in TKG-DM.

For the foreground, self-attention ensures internal consistency and coherence within the object, while cross-attention aligns the generated content with the text prompt. Due to the inherent bias in the training dataset \cite{schuhmann2022laion}, where foreground objects are more prominently described in captions, cross-attention forms a strong link between the foreground and the text prompt.

For the background, the init color noise introduced by channel mean shift dominates the generation process. 
The self-attention mechanism synergizes with this modified initial noise to guide the background toward the specified chroma key color. 
Since background elements are often less detailed or vaguely described in dataset captions, the cross-attention has a weaker influence on the background, allowing the init color noise to take precedence.

By exploiting this bias and manipulating the initial noise, TKG-DM effectively decouples the background from the text prompt. This results in a uniform chroma key background and enables the isolated generation of foreground content without interference from undesired background details.

\begin{figure*}[t]
    \centering
    \includegraphics[width=\textwidth]{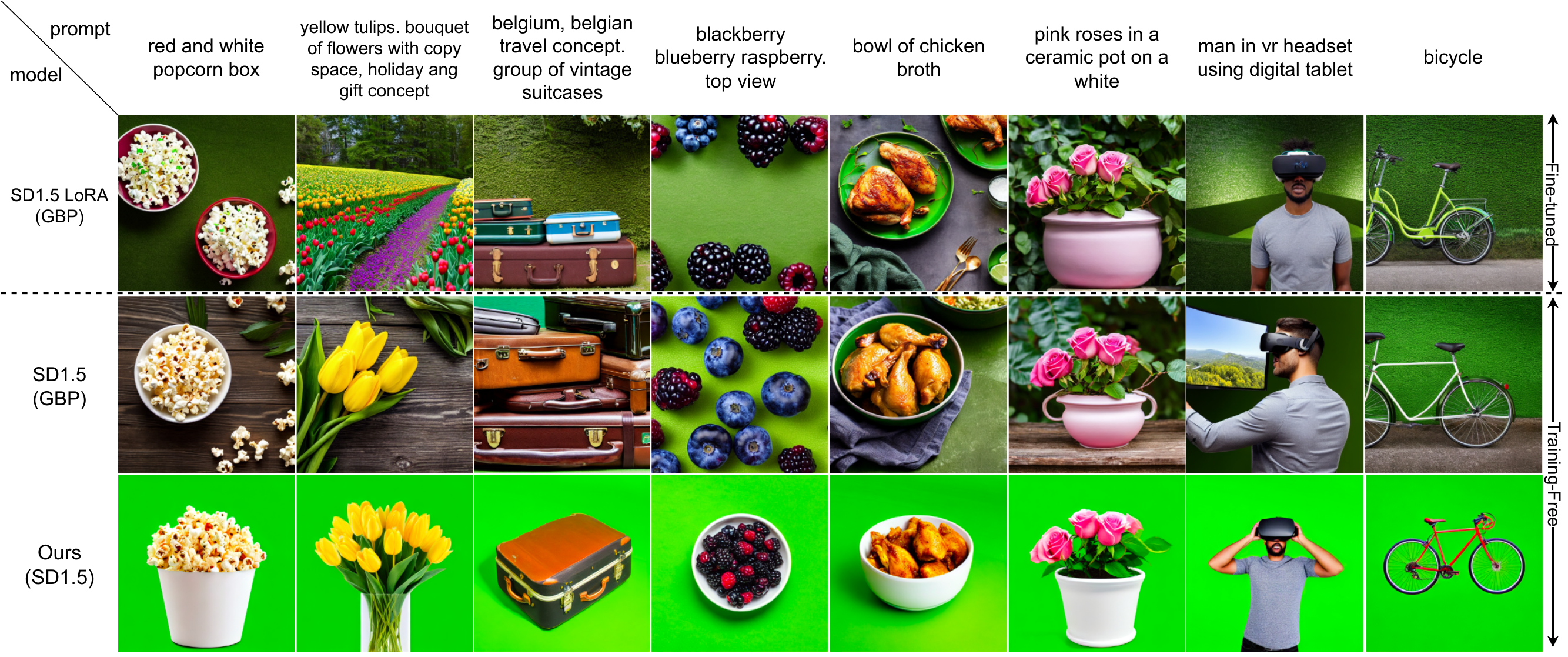} 
    \caption{\textbf{Qualitative comparison in SD1.5.} Existing methods fail to produce an accurate chroma key background. In contrast, ours produces a highly accurate chroma key background while generating high-quality foregrounds without Green Background Prompt (GBP).}
    \label{fig:result15}
\end{figure*}

\section{Experiments}
\label{sec:experiments}
\subsection{Experimental Setup}
We adopt SD1.5 and SDXL in TKG-DM to generate images at $512\times512$ and $1024\times1024$ resolutions, respectively, with DDIMScheduler, a guidance scale of 7.5, and 50 denoising steps. To obtain the init color noise for a green background, we apply channel mean shift with channels 2 and 3 in the positive direction for $\text{TargetShift}_c = +7\%$. Gaussian mask parameters are set to $(\sigma, \mu_i, \mu_j) = (0.5, w/2, h/2)$.

We compare our method against the naive usage of SD1.5 \cite{rombach2022high}, SDXL \cite{podell2023sdxl}, and DeepFloyd \cite{saharia2022photorealistic} using a Green Background Prompt (GBP) (=``isolated on a solid green background'') following the setup of MAGICK \cite{burgert2024magick}, as well as against fine-tuned models, namely GreenBack LoRA from CivitAI\footnote{\url{https://civitai.com/}} and LayerDiffuse \cite{zhang2024transparent}.

\begin{figure*}[t]
    \centering
    \includegraphics[width=\textwidth]{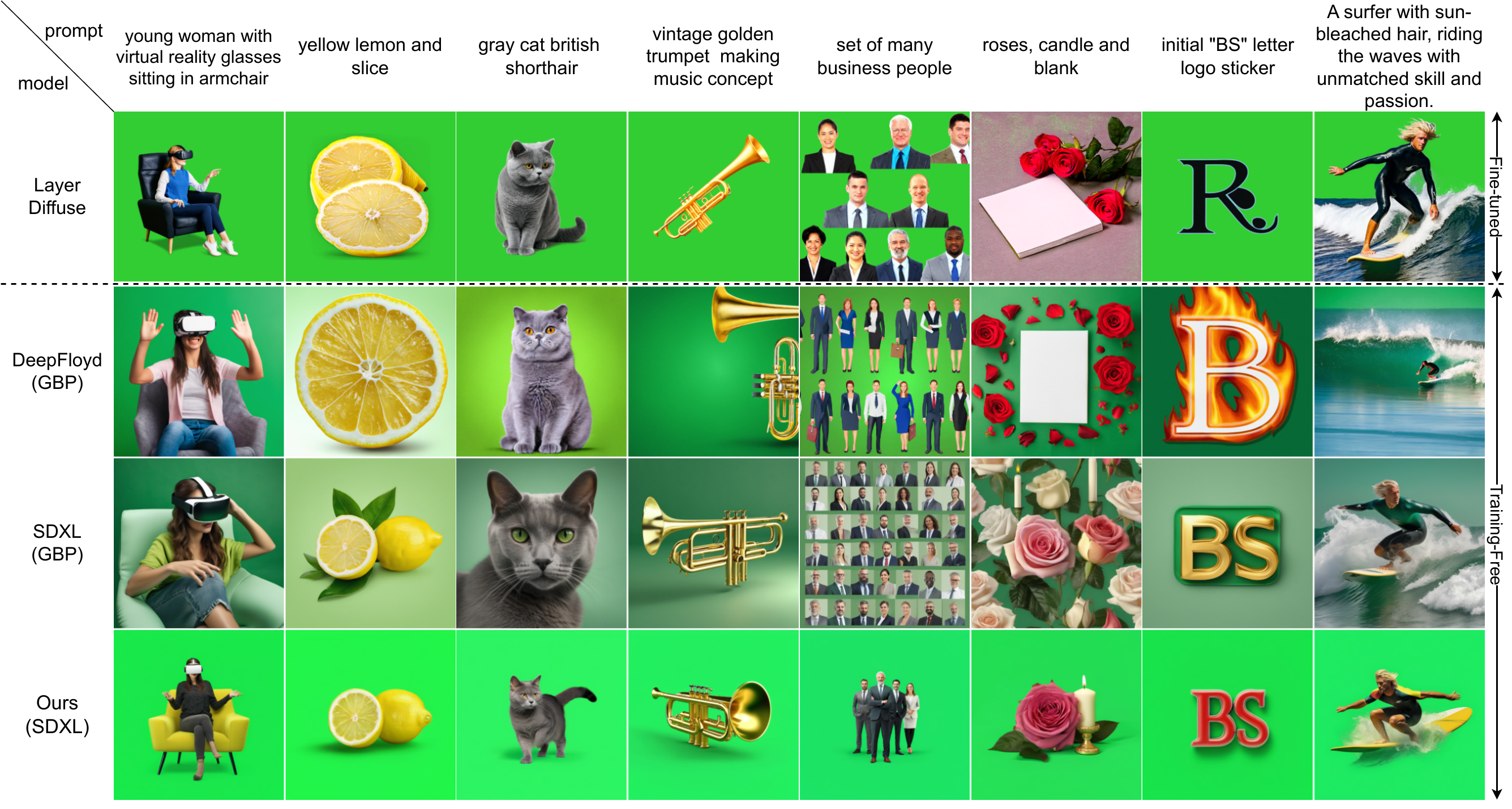} 
    \caption{\textbf{Qualitative comparison in SDXL.} Existing training-free methods fail to isolate foreground objects from the background. While LayerDiffuse, a fine-tuned model, successfully generates the transparent foregrounds on the lime green background, it may struggle with accurate text generation and background handling. In contrast, our model efficiently generates highly accurate chroma key images with foreground content placed over the background without fine-tuning and needing Green Background Prompt (GBP).}
    \label{fig:resultxl}
\end{figure*}
\begin{figure}[t]
    \centering
    \includegraphics[width=0.47\textwidth]{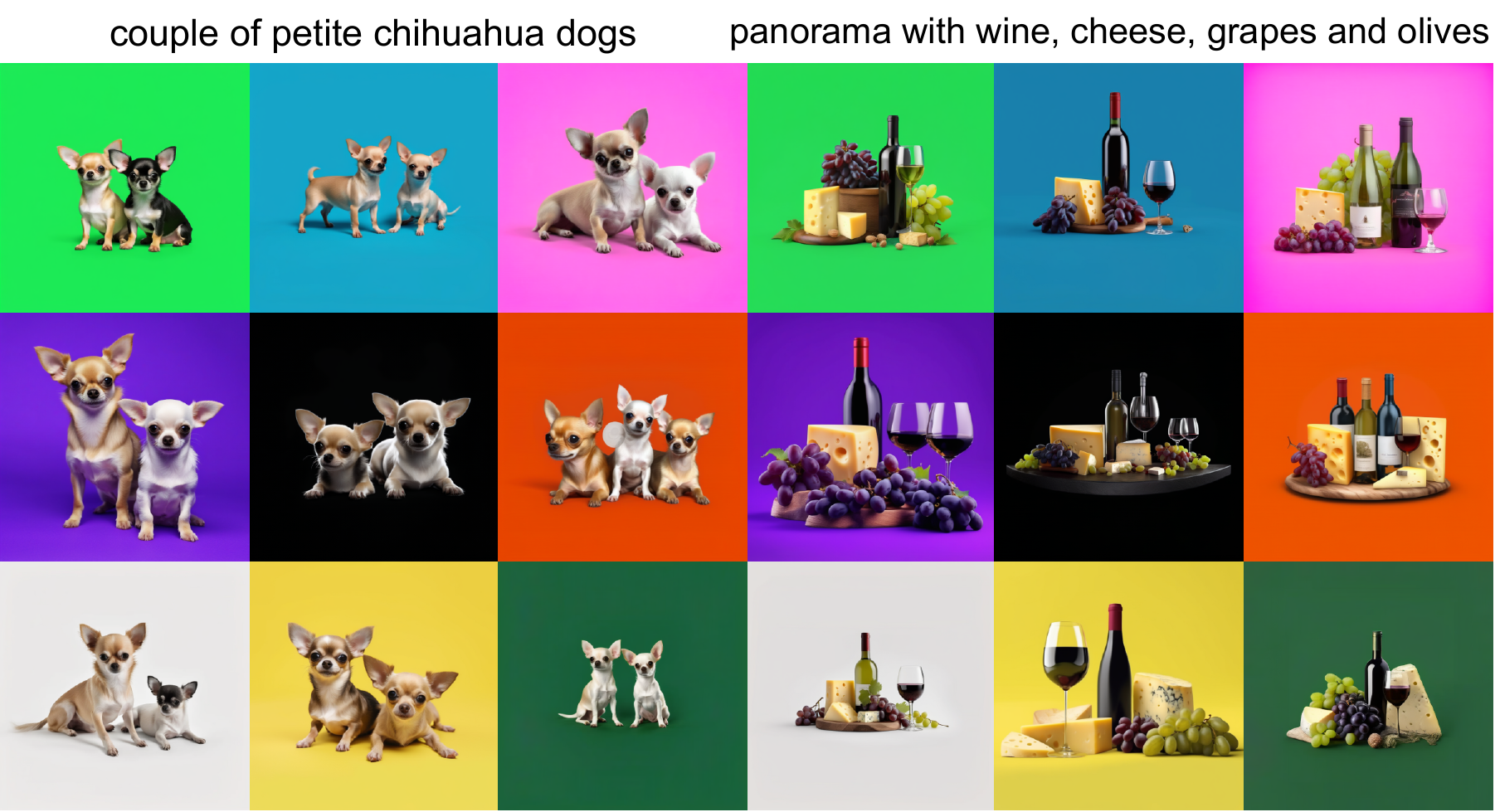} 
    \caption{\textbf{Qualitative various color result in SDXL.}
     By adjusting the channel mean shift, TKG-DM controls the background color while generating a high-quality foreground.}
    \label{fig:result_color}
\end{figure}
\subsection{Dataset and Metrics}
We collect 3,000 images from the MAGICK dataset \cite{burgert2024magick}, tailored for foreground image generation from text prompts. 
The MAGICK dataset is constructed by first generating monochromatic backgrounds using Deepfloyd~\cite{saharia2022photorealistic} with GBP, followed by quality enhancement using SDEdit \cite{meng2022sdedit} and SDXL’s img2img \cite{podell2023sdxl}. 
Foreground images with alpha values are curated using pixel-, deep learning-, and human-based methods. 
Our experiment utilizes alpha images, alpha masks, and corresponding text prompts.

We construct ground truth images for evaluation by overlaying the alpha images onto a lime green background images (RGB = 50, 205, 50).
We use the Fréchet Inception Distance (FID) to assess foreground quality \cite{heusel2017gans}. Additionally, we introduce a novel m-FID metric, which compares the mask extracted from generated chroma key content using BiRefNet \cite{zheng2024bilateral} to the ground truth mask, providing a comprehensive assessment of foreground accuracy.
For semantic alignment, we employ CLIP-Sentence (CLIP-S) and CLIP-Image (CLIP-I) \cite{hessel2021clipscore}: CLIP-S measures alignment with the text, while CLIP-I assesses visual similarity to the ground truth.

\subsection{Qualitative Result}
\label{sec:qualitative}
Fig.~\ref{fig:result15} and Fig.~\ref{fig:resultxl} show the qualitative results of our model and existing methods with SD1.5 and SDXL, respectively. 
They demonstrate that our model generates high-precision chroma key images without prompt engineering.

With the Green Background Prompt (GBP), SD1.5 struggles to generate a clean green background, while SDXL, while slightly better, generates unstable light green tints on the background and often produces multiple foreground objects, complicating foreground-background separation.
Additionally, the GBP’s ``green`` element unintentionally adds green tones to the foreground, such as on clothing or in a cat’s eyes (Fig.~\ref{fig:resultxl}, first and third columns). DeepFloyd performs the worst foreground quality and text alignment.

In the fine-tuned model, while LoRA (SD1.5) shows minor improvements, it still struggles with chroma key accuracy. 
LayerDiffuse produces well-separated foregrounds but occasionally loses details, like precise numbers or letters, due to dataset limitations in fine-tuning. 
Mask generation also occasionally fails, resulting in uncut images.

Moreover, existing models struggle to generate accurate chroma key images when background information is included in the text prompt (Fig.~\ref{fig:resultxl}, last columns). 
In contrast, our training-free model consistently produces high-quality chroma key backgrounds with only the foreground content, ensuring precise separation. 

Furthermore, as shown in Fig.~\ref{fig:result_color}, TKG-DM generates the various color background images through channel mean shift without prompt engineering. Additional results are provided in the supplementary material.

\begin{table}[t]
  \caption{\textbf{Quantitative results.} It indicates that our training-free model outperforms the existing model with Green Background Prompt (GBP) and rivals the fine-tuned model, LayerDiffuse \cite{zhang2024transparent}.
  }
  \label{table:qualitative}
  \centering
  \setlength{\tabcolsep}{3mm}
  \resizebox{\columnwidth}{!}{%
  \begin{tabular}{l|cccc}
    \hline
    \multicolumn{5}{c}{\textbf{SD1.5}}\\
    \hline
     & FID  $\downarrow$ & m-FID $\downarrow$ & CLIP-I$\uparrow$ & CLIP-S $\uparrow$ \\
    \hline
    SD1.5 (GBP) \cite{rombach2022high}   & 85.00 & 63.54 & \underline{0.710} & \underline{0.256}\\
    GB LoRA (GBP) & \underline{60.29} & \underline{49.03} & 0.704 & 0.243\\
    \textbf{Ours}                & \textbf{56.32} & \textbf{40.75}  & \textbf{0.737} & \textbf{0.261}\\
    \hline
    \multicolumn{5}{c}{\textbf{SDXL}}\\
    \hline
    DeepFloyd (GBP) \cite{saharia2022photorealistic} &  31.57 & 20.31 & 0.781  & 0.270\\
    \hline
    SDXL (GBP) \cite{podell2023sdxl} & 45.32 & 39.17   & 0.759 & 0.272\\
    LayerDiffuse \cite{zhang2024transparent}     & \textbf{29.34}   & \textbf{29.82}   & \textbf{0.778}   & \textbf{0.276}\\
    \textbf{Ours}                & \underline{41.81} & \underline{31.43} & \underline{0.763} & \underline{0.273}\\
    
    \hline
    
  \end{tabular}
  }
\end{table}
\begin{figure}[t]
    \centering
    \includegraphics[width=0.47\textwidth]{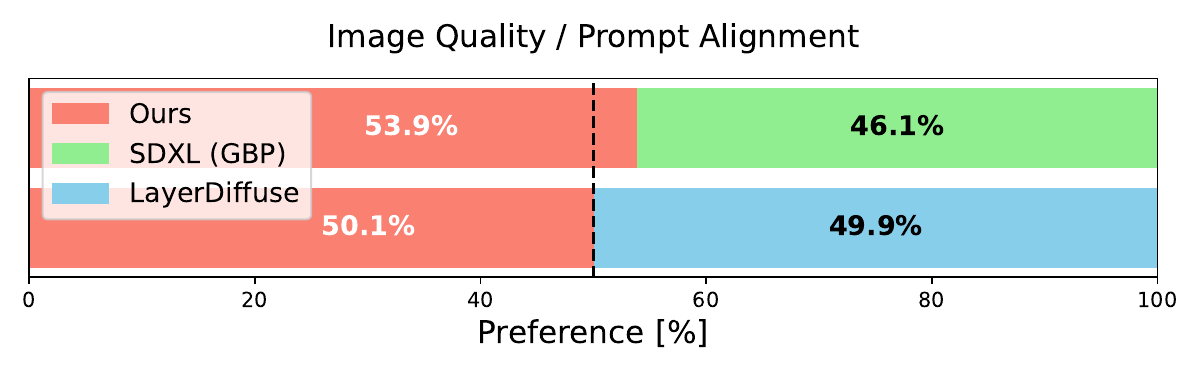}
    \caption{Results from our user study show the preference rates for foreground objects generated by our model and existing models based on their image quality and alignment with the text prompt. GBP means a model with a Green Background Prompt.}
    \label{fig:user_study}
\end{figure}
\begin{figure*}[t]
    \centering
    \includegraphics[width=\textwidth]{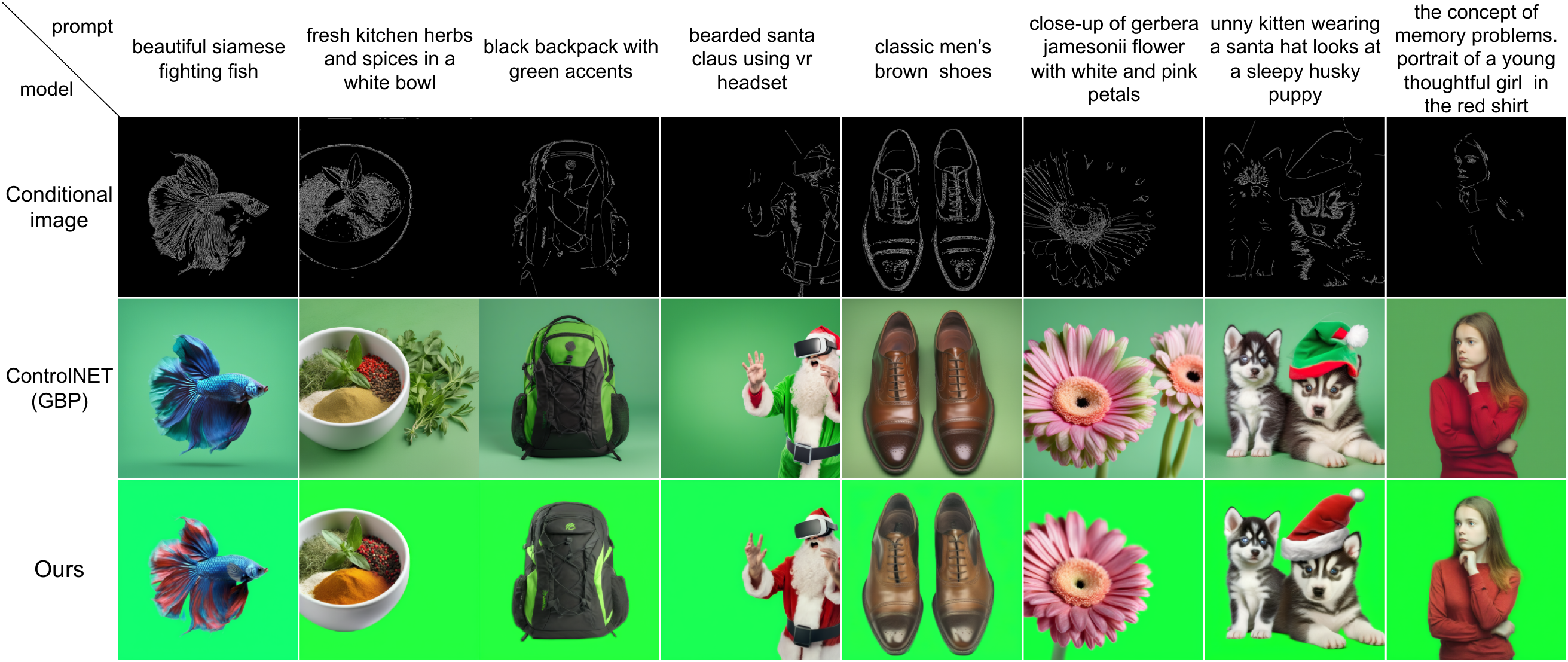} 
    \caption{\textbf{Qualitative comparison in ControlNET.} Existing methods typically generate images with issues such as color erosion in the foreground, uneven backgrounds, or unintended elements from the conditioning input. In contrast, our model generates images with the foreground isolated from the background, maintaining a consistent chroma key background without the Green Background Prompt (GBP).}
    \label{fig:result-controlnet}
\end{figure*}


\subsection{Quantitative Result}
As shown in Table \ref{table:qualitative}, our method consistently outperforms other training-free models across all metrics, achieving better image realism and mask accuracy, as indicated by lower FID and m-FID scores. It also aligns well with input text prompts and has closer visual similarity to the ground truth, as indicated by higher CLIP-S and CLIP-I scores.

In the SDXL setting, LayerDiffuse achieves the best FID, while our model remains competitive across all other metrics. The FID advantage for LayerDiffuse likely results from its setup, which overlays foregrounds on a lime green background matching the ground truth.

DeepFloyd's high FID, m-FID, and CLIP-I scores reflect its similarity to the ground truth based on DeepFloyd's outputs. However, this alignment gives it an inherent advantage, making it unsuitable as a fair benchmark for image quality. Its lower CLIP-S score further indicates weaker text alignment compared to other models.

Overall, these results underscore our model’s ability to generate high-quality, text-aligned foregrounds without fine-tuning, offering an efficient chroma key content generation solution.


\subsection{User Study}

Besides objective metrics, we conducted a user study to assess subjective human preferences for prompt adherence and image quality, comparing our method against SDXL with Green Background Prompt (GBP) and LayerDiffuse. 
We generated 30 image pairs per method, totaling 60 images. 
100 participants ranked images via two alternative forced-choice~\cite{saharia2022image, moser2023yoda} and based on foreground quality and text alignment. 
To ensure focus on the foreground, we used a foreground extraction model~\cite{zheng2024bilateral} and manually refined areas as needed. 
As shown in Fig.~\ref{fig:user_study}, our training-free approach improves object realism and text alignment over the existing model.

\begin{table}[t]
  \caption{Quantitative results comparing our method with existing methods for ControlNet~\cite{zhang2023adding}. GBP means a model with a Green Background Prompt.
  }
  \label{table:controlnet}
  \centering
  \setlength{\tabcolsep}{3mm}
  \resizebox{\columnwidth}{!}{%
  \begin{tabular}{l|cccc}
    \hline
    \multicolumn{5}{c}{\textbf{ControlNet} \cite{zhang2023adding}} \\
    \hline
     & FID $\downarrow$ & m-FID $\downarrow$ & CLIP-I  $\uparrow$ & CLIP-S $\uparrow$ \\
    \hline
    SDXL (GBP) \cite{podell2023sdxl}   & 22.04 & 18.62 & 0.819 & 0.279 \\
    \textbf{Ours}  & \textbf{17.09} & \textbf{17.22} & \textbf{0.834} & \textbf{0.284} \\
    \hline
  \end{tabular}%
  }
\end{table}

\section{Applications}
\label{sec:application}
We demonstrate the flexibility of TKG-DM in various tasks, implemented with the SD1.5 and SDXL, as presented in Table~\ref{table:controlnet}, Fig.~\ref{fig:result-controlnet} and Fig.~\ref{fig:application}. Further details and additional results are provided in the supplementary material.

\textbf{Application to ControlNet.}
As depicted in Fig. \ref{fig:result-controlnet} and Table \ref{table:controlnet}, our method integrates seamlessly with ControlNet \cite{zhang2023adding}, enabling precise generation control using conditioning inputs like canny edges. By applying a Gaussian mask to the init color noise, we preserve normal initial noise in the foreground regions, resulting in distinct foregrounds on uniform chroma key backgrounds.

Compared to the standard ControlNet, our method demonstrates superior qualitative and quantitative performance. Existing methods often face issues like color erosion in the foreground, uneven backgrounds, or unintended elements from the conditioning input. By isolating the foreground from background influence and maintaining a consistent chroma key background, our approach is highly effective for compositing tasks without prompt engineering.

\textbf{Layout-Aware Text-to-Image Generation.}
As discussed in Section~\ref{sec:init_selector}, our model enables control over foreground objects' location and size by adjusting the Gaussian mask's center and spread. This control allows users to place objects precisely within specific areas of the generated image, making it practical for layout-aware tasks. Additionally, using multiple Gaussian masks enables the generation of multiple foreground objects within a single scene (Fig. \ref{fig:application}, top-right). Unlike existing models that lack targeted control over object placement and size, our approach integrates layout information seamlessly without fine-tuning.

\begin{figure}[t]
    \centering
    \includegraphics[width=0.47\textwidth]{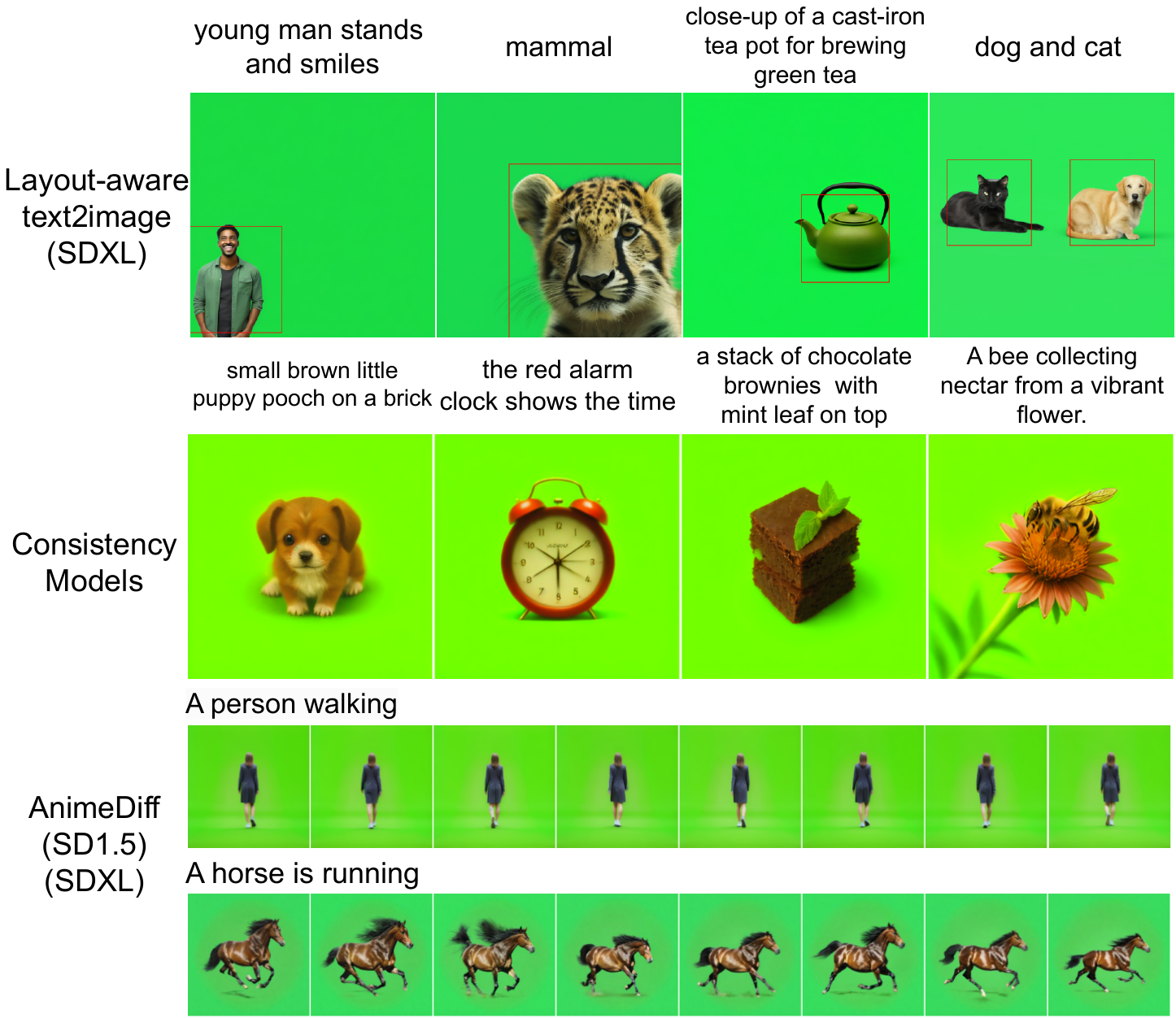} 
    \caption{\textbf{Application results of TKG-DM across various tasks.} TKG-DM effectively supports layout-aware text-to-image generation, consistency models and text-to-video. Each row demonstrates its versatility in adapting to different domains, from realistic object and character placements to animation.}
    \label{fig:application}
\end{figure}

\textbf{Application to Consistency Models.}
Consistency Models \cite{song2023consistency} are generative models that produce high-quality outputs in a few steps, unlike traditional diffusion models. Operating in latent space, Consistency Models demonstrate the robustness of TKG-DM to generative models that handle images in the latent diffusion-based model, enabling significantly faster generation while maintaining high output quality without modifications.

\textbf{Application to Text-to-Video.}
Our TKG-DM extends seamlessly to text-to-video tasks, ensuring consistent foreground content across frames. We align foreground objects with the input prompt by applying TKG-DM with AnimateDiff \cite{guo2023animatediff} on each frame while maintaining a uniform background for chroma key video generation. This approach, the first to generate foreground video with controlled chroma key backgrounds, facilitates efficient and flexible video editing workflows involving background removal or replacement.

\section{Limitation and Future Work}
\label{sec:future}
\begin{figure}[t]
    \centering
    \includegraphics[width=0.47\textwidth]{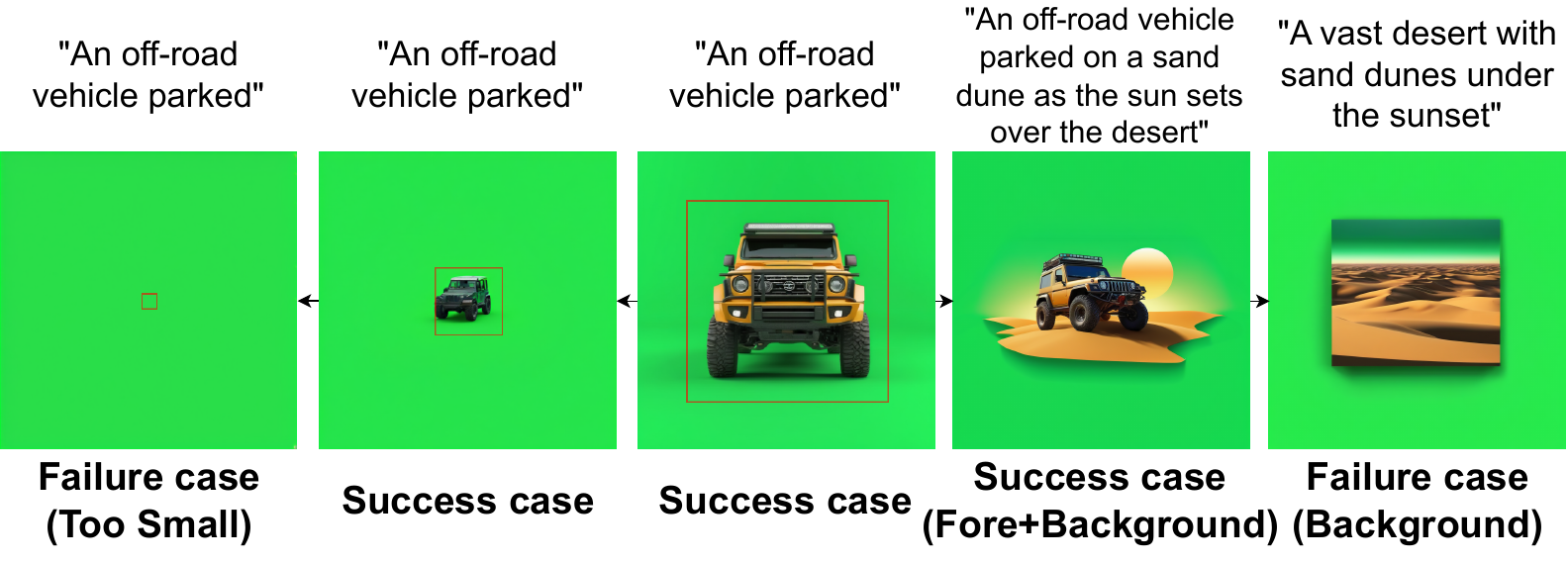} 
    \caption{\textbf{Failure Cases.} TKG-DM fails to generate a foreground object when the size is too small or when the text prompt lacks object information, resulting in only the background being generated. }
    \label{fig:fail}
\end{figure}

As shown in Fig. \ref{fig:fail}, TKG-DM primarily focuses on generating foreground content (e.g., objects and people), which limits its ability to generate backgrounds like landscapes. 
Additionally, if the size parameter is too small, ignore generating the foreground object \cite{shirakawa2024noisecollage, mao2023semantic}.

Future enhancements could expand TKG-DM’s capabilities to include controlled background generation, allowing separate and adjustable generation of both foreground and background content.
This would greatly benefit applications in video and 3D content creation where isolated and reusable background assets are essential.
\section{Conclusion}
\label{sec:conclusion}
This work presents the first exploration of initial noise color manipulation in diffusion models, demonstrating its effectiveness in controlling background chromaticity. Leveraging this insight, we introduce the Training-Free Chroma Key Content Generation Diffusion Model (TKG-DM), which generates prompt-based foregrounds on a chroma key background without fine-tuning. TKG-DM extends seamlessly to tasks like consistency models and text-to-video, highlighting its versatility beyond static images. Experimental results show that our method matches or surpasses fine-tuned models in both qualitative and quantitative evaluations without additional training. This approach not only advances theoretical understanding but also provides practical benefits for real-world applications where isolating foreground content is essential.

\section*{Acknowledgements}
This research was supported by BMBF project SustainML (Grant 101070408) and the BMDV NITD (45KI22B021).

{
    \small
    \bibliographystyle{ieeenat_fullname}
    \bibliography{main}

\begin{thebibliography}{63}
\providecommand{\natexlab}[1]{#1}
\providecommand{\url}[1]{\texttt{#1}}
\expandafter\ifx\csname urlstyle\endcsname\relax
  \providecommand{\doi}[1]{doi: #1}\else
  \providecommand{\doi}{doi: \begingroup \urlstyle{rm}\Url}\fi

\bibitem[Ban et~al.(2024)Ban, Wang, Zhou, Gong, Hsieh, and Cheng]{ban2024crystal}
Yuanhao Ban, Ruochen Wang, Tianyi Zhou, Boqing Gong, Cho-Jui Hsieh, and Minhao Cheng.
\newblock The crystal ball hypothesis in diffusion models: Anticipating object positions from initial noise.
\newblock \emph{arXiv preprint arXiv:2406.01970}, 2024.

\bibitem[Bar-Tal et~al.(2023)Bar-Tal, Yariv, Lipman, and Dekel]{bartal2023multidiffusionfusingdiffusionpaths}
Omer Bar-Tal, Lior Yariv, Yaron Lipman, and Tali Dekel.
\newblock Multidiffusion: Fusing diffusion paths for controlled image generation, 2023.

\bibitem[Benny and Wolf(2020)]{benny2020onegan}
Yaniv Benny and Lior Wolf.
\newblock Onegan: Simultaneous unsupervised learning of conditional image generation, foreground segmentation, and fine-grained clustering.
\newblock In \emph{ECCV}, pages 514--530. Springer, 2020.

\bibitem[Brooks et~al.(2023)Brooks, Holynski, and Efros]{brooks2023instructpix2pix}
Tim Brooks, Aleksander Holynski, and Alexei~A Efros.
\newblock Instructpix2pix: Learning to follow image editing instructions.
\newblock In \emph{CVPR}, pages 18392--18402, 2023.

\bibitem[Burgert et~al.(2024)Burgert, Price, Kuen, Li, and Ryoo]{burgert2024magick}
Ryan~D Burgert, Brian~L Price, Jason Kuen, Yijun Li, and Michael~S Ryoo.
\newblock Magick: A large-scale captioned dataset from matting generated images using chroma keying.
\newblock In \emph{CVPR}, pages 22595--22604, 2024.

\bibitem[Cao et~al.(2023)Cao, Wang, Qi, Shan, Qie, and Zheng]{cao2023masactrl}
Mingdeng Cao, Xintao Wang, Zhongang Qi, Ying Shan, Xiaohu Qie, and Yinqiang Zheng.
\newblock Masactrl: Tuning-free mutual self-attention control for consistent image synthesis and editing.
\newblock In \emph{ICCV}, pages 22560--22570, 2023.

\bibitem[Chai et~al.(2023)Chai, Zhuang, and Yan]{chai2023layoutdm}
Shang Chai, Liansheng Zhuang, and Fengying Yan.
\newblock Layoutdm: Transformer-based diffusion model for layout generation.
\newblock In \emph{CVPR}, pages 18349--18358, 2023.

\bibitem[Chefer et~al.(2023)Chefer, Alaluf, Vinker, Wolf, and Cohen-Or]{chefer2023attend}
Hila Chefer, Yuval Alaluf, Yael Vinker, Lior Wolf, and Daniel Cohen-Or.
\newblock Attend-and-excite: Attention-based semantic guidance for text-to-image diffusion models.
\newblock \emph{ACM Transactions on Graphics (TOG)}, 42\penalty0 (4):\penalty0 1--10, 2023.

\bibitem[Chen et~al.(2024)Chen, Wang, Nie, Zhang, and Liu]{chen2024anyscene}
Ruidong Chen, Lanjun Wang, Weizhi Nie, Yongdong Zhang, and An-An Liu.
\newblock Anyscene: Customized image synthesis with composited foreground.
\newblock In \emph{CVPR}, pages 8724--8733, 2024.

\bibitem[Choi et~al.(2024)Choi, Kwak, Lee, Choi, and Shin]{choi2024improving}
Yisol Choi, Sangkyung Kwak, Kyungmin Lee, Hyungwon Choi, and Jinwoo Shin.
\newblock Improving diffusion models for virtual try-on.
\newblock \emph{arXiv preprint arXiv:2403.05139}, 2024.

\bibitem[Dalva et~al.(2024)Dalva, Li, Liu, Zhao, Zhang, Lin, and Yanardag]{dalva2024layerfusion}
Yusuf Dalva, Yijun Li, Qing Liu, Nanxuan Zhao, Jianming Zhang, Zhe Lin, and Pinar Yanardag.
\newblock Layerfusion: Harmonized multi-layer text-to-image generation with generative priors.
\newblock \emph{arXiv preprint arXiv:2412.04460}, 2024.

\bibitem[Dhariwal and Nichol(2021)]{dhariwal2021diffusion}
Prafulla Dhariwal and Alexander Nichol.
\newblock Diffusion models beat gans on image synthesis.
\newblock \emph{NeurIPS}, 34:\penalty0 8780--8794, 2021.

\bibitem[Eyring et~al.(2024)Eyring, Karthik, Roth, Dosovitskiy, and Akata]{eyring2024reno}
Luca Eyring, Shyamgopal Karthik, Karsten Roth, Alexey Dosovitskiy, and Zeynep Akata.
\newblock Reno: Enhancing one-step text-to-image models through reward-based noise optimization.
\newblock \emph{arXiv preprint arXiv:2406.04312}, 2024.

\bibitem[Guo et~al.(2024)Guo, Liu, Cui, Li, Yang, and Huang]{guo2024initno}
Xiefan Guo, Jinlin Liu, Miaomiao Cui, Jiankai Li, Hongyu Yang, and Di Huang.
\newblock Initno: Boosting text-to-image diffusion models via initial noise optimization.
\newblock In \emph{CVPR}, pages 9380--9389, 2024.

\bibitem[Guo et~al.(2023)Guo, Yang, Rao, Liang, Wang, Qiao, Agrawala, Lin, and Dai]{guo2023animatediff}
Yuwei Guo, Ceyuan Yang, Anyi Rao, Zhengyang Liang, Yaohui Wang, Yu Qiao, Maneesh Agrawala, Dahua Lin, and Bo Dai.
\newblock Animatediff: Animate your personalized text-to-image diffusion models without specific tuning.
\newblock \emph{arXiv preprint arXiv:2307.04725}, 2023.

\bibitem[Hertz et~al.(2022)Hertz, Mokady, Tenenbaum, Aberman, Pritch, and Cohen-Or]{hertz2022prompt}
Amir Hertz, Ron Mokady, Jay Tenenbaum, Kfir Aberman, Yael Pritch, and Daniel Cohen-Or.
\newblock Prompt-to-prompt image editing with cross attention control.
\newblock \emph{arXiv preprint arXiv:2208.01626}, 2022.

\bibitem[Hessel et~al.(2021)Hessel, Holtzman, Forbes, Bras, and Choi]{hessel2021clipscore}
Jack Hessel, Ari Holtzman, Maxwell Forbes, Ronan~Le Bras, and Yejin Choi.
\newblock Clipscore: A reference-free evaluation metric for image captioning.
\newblock \emph{arXiv preprint arXiv:2104.08718}, 2021.

\bibitem[Heusel et~al.(2017)Heusel, Ramsauer, Unterthiner, Nessler, and Hochreiter]{heusel2017gans}
Martin Heusel, Hubert Ramsauer, Thomas Unterthiner, Bernhard Nessler, and Sepp Hochreiter.
\newblock Gans trained by a two time-scale update rule converge to a local nash equilibrium.
\newblock \emph{NeurIPS}, 30, 2017.

\bibitem[Ho and Salimans(2022)]{ho2022classifier}
Jonathan Ho and Tim Salimans.
\newblock Classifier-free diffusion guidance.
\newblock \emph{arXiv preprint arXiv:2207.12598}, 2022.

\bibitem[Ho et~al.(2020)Ho, Jain, and Abbeel]{ho2020denoising}
Jonathan Ho, Ajay Jain, and Pieter Abbeel.
\newblock Denoising diffusion probabilistic models.
\newblock \emph{NeurIPS}, 33:\penalty0 6840--6851, 2020.

\bibitem[Huang et~al.(2023)Huang, Huang, Yang, Ren, Liu, Li, Ye, Liu, Yin, and Zhao]{huang2023make}
Rongjie Huang, Jiawei Huang, Dongchao Yang, Yi Ren, Luping Liu, Mingze Li, Zhenhui Ye, Jinglin Liu, Xiang Yin, and Zhou Zhao.
\newblock Make-an-audio: Text-to-audio generation with prompt-enhanced diffusion models.
\newblock In \emph{ICML}, pages 13916--13932. PMLR, 2023.

\bibitem[Huang et~al.(2024)Huang, Cai, Han, Liang, Pei, Lu, Xu, Zhang, and Xu]{huang2024layerdiff}
Runhui Huang, Kaixin Cai, Jianhua Han, Xiaodan Liang, Renjing Pei, Guansong Lu, Songcen Xu, Wei Zhang, and Hang Xu.
\newblock Layerdiff: Exploring text-guided multi-layered composable image synthesis via layer-collaborative diffusion model.
\newblock In \emph{European Conference on Computer Vision}, pages 144--160. Springer, 2024.

\bibitem[Inoue et~al.(2023)Inoue, Kikuchi, Simo-Serra, Otani, and Yamaguchi]{inoue2023layoutdm}
Naoto Inoue, Kotaro Kikuchi, Edgar Simo-Serra, Mayu Otani, and Kota Yamaguchi.
\newblock Layoutdm: Discrete diffusion model for controllable layout generation.
\newblock In \emph{CVPR}, pages 10167--10176, 2023.

\bibitem[Kong et~al.(2020)Kong, Ping, Huang, Zhao, and Catanzaro]{kong2020diffwave}
Zhifeng Kong, Wei Ping, Jiaji Huang, Kexin Zhao, and Bryan Catanzaro.
\newblock Diffwave: A versatile diffusion model for audio synthesis.
\newblock \emph{arXiv preprint arXiv:2009.09761}, 2020.

\bibitem[Lipman et~al.(2022)Lipman, Chen, Ben-Hamu, Nickel, and Le]{lipman2022flow}
Yaron Lipman, Ricky~TQ Chen, Heli Ben-Hamu, Maximilian Nickel, and Matt Le.
\newblock Flow matching for generative modeling.
\newblock \emph{arXiv preprint arXiv:2210.02747}, 2022.

\bibitem[Lugmayr et~al.(2022)Lugmayr, Danelljan, Romero, Yu, Timofte, and Van~Gool]{lugmayr2022repaint}
Andreas Lugmayr, Martin Danelljan, Andres Romero, Fisher Yu, Radu Timofte, and Luc Van~Gool.
\newblock Repaint: Inpainting using denoising diffusion probabilistic models.
\newblock In \emph{CVPR}, pages 11461--11471, 2022.

\bibitem[Mao et~al.(2023{\natexlab{a}})Mao, Wang, and Aizawa]{mao2023guided}
Jiafeng Mao, Xueting Wang, and Kiyoharu Aizawa.
\newblock Guided image synthesis via initial image editing in diffusion model.
\newblock \emph{arXiv preprint arXiv:2305.03382}, 2023{\natexlab{a}}.

\bibitem[Mao et~al.(2023{\natexlab{b}})Mao, Wang, and Aizawa]{mao2023semantic}
Jiafeng Mao, Xueting Wang, and Kiyoharu Aizawa.
\newblock Semantic-driven initial image construction for guided image synthesis in diffusion model.
\newblock \emph{arXiv preprint arXiv:2312.08872}, 2023{\natexlab{b}}.

\bibitem[Meng et~al.(2022)Meng, He, Song, Song, Wu, Zhu, and Ermon]{meng2022sdedit}
Chenlin Meng, Yutong He, Yang Song, Jiaming Song, Jiajun Wu, Jun-Yan Zhu, and Stefano Ermon.
\newblock {SDE}dit: Guided image synthesis and editing with stochastic differential equations.
\newblock In \emph{ICLR}, 2022.

\bibitem[Morita et~al.(2023{\natexlab{a}})Morita, Zhang, Ho, and Zhou]{morita2023interactive}
Ryugo Morita, Zhiqiang Zhang, Man~M Ho, and Jinjia Zhou.
\newblock Interactive image manipulation with complex text instructions.
\newblock In \emph{WACV}, pages 1053--1062, 2023{\natexlab{a}}.

\bibitem[Morita et~al.(2023{\natexlab{b}})Morita, Zhang, and Zhou]{morita2023batinet}
Ryugo Morita, Zhiqiang Zhang, and Jinjia Zhou.
\newblock Batinet: Background-aware text to image synthesis and manipulation network.
\newblock In \emph{ICIP}, pages 765--769. IEEE, 2023{\natexlab{b}}.

\bibitem[Moser et~al.(2023)Moser, Frolov, Raue, Palacio, and Dengel]{moser2023yoda}
Brian~B Moser, Stanislav Frolov, Federico Raue, Sebastian Palacio, and Andreas Dengel.
\newblock Yoda: You only diffuse areas. an area-masked diffusion approach for image super-resolution.
\newblock \emph{arXiv preprint arXiv:2308.07977}, 2023.

\bibitem[Moser et~al.(2024)Moser, Shanbhag, Raue, Frolov, Palacio, and Dengel]{moser2024diffusion}
Brian~B Moser, Arundhati~S Shanbhag, Federico Raue, Stanislav Frolov, Sebastian Palacio, and Andreas Dengel.
\newblock Diffusion models, image super-resolution, and everything: A survey.
\newblock \emph{IEEE Transactions on Neural Networks and Learning Systems}, 2024.

\bibitem[Nichol et~al.(2021)Nichol, Dhariwal, Ramesh, Shyam, Mishkin, McGrew, Sutskever, and Chen]{nichol2021glide}
Alex Nichol, Prafulla Dhariwal, Aditya Ramesh, Pranav Shyam, Pamela Mishkin, Bob McGrew, Ilya Sutskever, and Mark Chen.
\newblock Glide: Towards photorealistic image generation and editing with text-guided diffusion models.
\newblock \emph{arXiv preprint arXiv:2112.10741}, 2021.

\bibitem[Nichol and Dhariwal(2021)]{nichol2021improved}
Alexander~Quinn Nichol and Prafulla Dhariwal.
\newblock Improved denoising diffusion probabilistic models.
\newblock In \emph{ICML}, pages 8162--8171. PMLR, 2021.

\bibitem[Podell et~al.(2023)Podell, English, Lacey, Blattmann, Dockhorn, M{\"u}ller, Penna, and Rombach]{podell2023sdxl}
Dustin Podell, Zion English, Kyle Lacey, Andreas Blattmann, Tim Dockhorn, Jonas M{\"u}ller, Joe Penna, and Robin Rombach.
\newblock Sdxl: Improving latent diffusion models for high-resolution image synthesis.
\newblock \emph{arXiv preprint arXiv:2307.01952}, 2023.

\bibitem[Qian et~al.(2023)Qian, Mai, Hamdi, Ren, Siarohin, Li, Lee, Skorokhodov, Wonka, Tulyakov, et~al.]{qian2023magic123}
Guocheng Qian, Jinjie Mai, Abdullah Hamdi, Jian Ren, Aliaksandr Siarohin, Bing Li, Hsin-Ying Lee, Ivan Skorokhodov, Peter Wonka, Sergey Tulyakov, et~al.
\newblock Magic123: One image to high-quality 3d object generation using both 2d and 3d diffusion priors.
\newblock \emph{arXiv preprint arXiv:2306.17843}, 2023.

\bibitem[Quattrini et~al.(2025)Quattrini, Pippi, Cascianelli, and Cucchiara]{quattrini2025alfie}
Fabio Quattrini, Vittorio Pippi, Silvia Cascianelli, and Rita Cucchiara.
\newblock Alfie: Democratising rgba image generation with no \$\$\$.
\newblock In \emph{European Conference on Computer Vision}, pages 38--55. Springer, 2025.

\bibitem[Ramesh et~al.(2021)Ramesh, Pavlov, Goh, Gray, Voss, Radford, Chen, and Sutskever]{ramesh2021zero}
Aditya Ramesh, Mikhail Pavlov, Gabriel Goh, Scott Gray, Chelsea Voss, Alec Radford, Mark Chen, and Ilya Sutskever.
\newblock Zero-shot text-to-image generation.
\newblock In \emph{ICML}, pages 8821--8831. Pmlr, 2021.

\bibitem[Rombach et~al.(2022)Rombach, Blattmann, Lorenz, Esser, and Ommer]{rombach2022high}
Robin Rombach, Andreas Blattmann, Dominik Lorenz, Patrick Esser, and Bj{\"o}rn Ommer.
\newblock High-resolution image synthesis with latent diffusion models.
\newblock In \emph{CVPR}, pages 10684--10695, 2022.

\bibitem[Saharia et~al.(2022{\natexlab{a}})Saharia, Chan, Saxena, Li, Whang, Denton, Ghasemipour, Gontijo~Lopes, Karagol~Ayan, Salimans, et~al.]{saharia2022photorealistic}
Chitwan Saharia, William Chan, Saurabh Saxena, Lala Li, Jay Whang, Emily~L Denton, Kamyar Ghasemipour, Raphael Gontijo~Lopes, Burcu Karagol~Ayan, Tim Salimans, et~al.
\newblock Photorealistic text-to-image diffusion models with deep language understanding.
\newblock \emph{NeurIPS}, 35:\penalty0 36479--36494, 2022{\natexlab{a}}.

\bibitem[Saharia et~al.(2022{\natexlab{b}})Saharia, Ho, Chan, Salimans, Fleet, and Norouzi]{saharia2022image}
Chitwan Saharia, Jonathan Ho, William Chan, Tim Salimans, David~J Fleet, and Mohammad Norouzi.
\newblock Image super-resolution via iterative refinement.
\newblock \emph{IEEE TPAMI}, 45\penalty0 (4):\penalty0 4713--4726, 2022{\natexlab{b}}.

\bibitem[Samuel et~al.(2024{\natexlab{a}})Samuel, Ben-Ari, Darshan, Maron, and Chechik]{samuel2024norm}
Dvir Samuel, Rami Ben-Ari, Nir Darshan, Haggai Maron, and Gal Chechik.
\newblock Norm-guided latent space exploration for text-to-image generation.
\newblock \emph{NeurIPS}, 36, 2024{\natexlab{a}}.

\bibitem[Samuel et~al.(2024{\natexlab{b}})Samuel, Ben-Ari, Raviv, Darshan, and Chechik]{samuel2024generating}
Dvir Samuel, Rami Ben-Ari, Simon Raviv, Nir Darshan, and Gal Chechik.
\newblock Generating images of rare concepts using pre-trained diffusion models.
\newblock In \emph{AAAI}, pages 4695--4703, 2024{\natexlab{b}}.

\bibitem[Schuhmann et~al.(2022)Schuhmann, Beaumont, Vencu, Gordon, Wightman, Cherti, Coombes, Katta, Mullis, Wortsman, et~al.]{schuhmann2022laion}
Christoph Schuhmann, Romain Beaumont, Richard Vencu, Cade Gordon, Ross Wightman, Mehdi Cherti, Theo Coombes, Aarush Katta, Clayton Mullis, Mitchell Wortsman, et~al.
\newblock Laion-5b: An open large-scale dataset for training next generation image-text models.
\newblock \emph{Advances in Neural Information Processing Systems}, 35:\penalty0 25278--25294, 2022.

\bibitem[Shirakawa and Uchida(2024)]{shirakawa2024noisecollage}
Takahiro Shirakawa and Seiichi Uchida.
\newblock Noisecollage: A layout-aware text-to-image diffusion model based on noise cropping and merging.
\newblock In \emph{CVPR}, pages 8921--8930, 2024.

\bibitem[Sohl-Dickstein et~al.(2015)Sohl-Dickstein, Weiss, Maheswaranathan, and Ganguli]{sohl2015deep}
Jascha Sohl-Dickstein, Eric Weiss, Niru Maheswaranathan, and Surya Ganguli.
\newblock Deep unsupervised learning using nonequilibrium thermodynamics.
\newblock In \emph{ICML}, pages 2256--2265. PMLR, 2015.

\bibitem[Song et~al.(2020)Song, Meng, and Ermon]{song2020denoising}
Jiaming Song, Chenlin Meng, and Stefano Ermon.
\newblock Denoising diffusion implicit models.
\newblock \emph{arXiv preprint arXiv:2010.02502}, 2020.

\bibitem[Song et~al.(2023)Song, Dhariwal, Chen, and Sutskever]{song2023consistency}
Yang Song, Prafulla Dhariwal, Mark Chen, and Ilya Sutskever.
\newblock Consistency models.
\newblock \emph{arXiv preprint arXiv:2303.01469}, 2023.

\bibitem[Vass(2024)]{vass2023sdxl}
Timothy~Alexis Vass.
\newblock Explaining the sdxl latent space.
\newblock \url{https://huggingface.co/blog/TimothyAlexisVass/explaining-the-sdxl-latent-space}, 2024.

\bibitem[Voleti et~al.(2024)Voleti, Yao, Boss, Letts, Pankratz, Tochilkin, Laforte, Rombach, and Jampani]{voleti2024sv3d}
Vikram Voleti, Chun-Han Yao, Mark Boss, Adam Letts, David Pankratz, Dmitry Tochilkin, Christian Laforte, Robin Rombach, and Varun Jampani.
\newblock Sv3d: Novel multi-view synthesis and 3d generation from a single image using latent video diffusion.
\newblock \emph{arXiv preprint arXiv:2403.12008}, 2024.

\bibitem[Voynov et~al.(2023)Voynov, Aberman, and Cohen-Or]{voynov2023sketch}
Andrey Voynov, Kfir Aberman, and Daniel Cohen-Or.
\newblock Sketch-guided text-to-image diffusion models.
\newblock In \emph{ACM SIGGRAPH 2023 Conference Proceedings}, pages 1--11, 2023.

\bibitem[Wu et~al.(2023)Wu, Ge, Wang, Lei, Gu, Shi, Hsu, Shan, Qie, and Shou]{wu2023tune}
Jay~Zhangjie Wu, Yixiao Ge, Xintao Wang, Stan~Weixian Lei, Yuchao Gu, Yufei Shi, Wynne Hsu, Ying Shan, Xiaohu Qie, and Mike~Zheng Shou.
\newblock Tune-a-video: One-shot tuning of image diffusion models for text-to-video generation.
\newblock In \emph{ICCV}, pages 7623--7633, 2023.

\bibitem[Xu et~al.(2024{\natexlab{a}})Xu, Zhang, and Shi]{xu2024good}
Katherine Xu, Lingzhi Zhang, and Jianbo Shi.
\newblock Good seed makes a good crop: Discovering secret seeds in text-to-image diffusion models.
\newblock \emph{arXiv preprint arXiv:2405.14828}, 2024{\natexlab{a}}.

\bibitem[Xu et~al.(2024{\natexlab{b}})Xu, Gu, Chen, and Chen]{xu2024ootdiffusion}
Yuhao Xu, Tao Gu, Weifeng Chen, and Chengcai Chen.
\newblock Ootdiffusion: Outfitting fusion based latent diffusion for controllable virtual try-on.
\newblock \emph{arXiv preprint arXiv:2403.01779}, 2024{\natexlab{b}}.

\bibitem[Yang et~al.(2023{\natexlab{a}})Yang, Gu, Zhang, Zhang, Chen, Sun, Chen, and Wen]{yang2023paint}
Binxin Yang, Shuyang Gu, Bo Zhang, Ting Zhang, Xuejin Chen, Xiaoyan Sun, Dong Chen, and Fang Wen.
\newblock Paint by example: Exemplar-based image editing with diffusion models.
\newblock In \emph{CVPR}, pages 18381--18391, 2023{\natexlab{a}}.

\bibitem[Yang et~al.(2023{\natexlab{b}})Yang, Wang, Gan, Li, Lin, Wu, Duan, Liu, Liu, Zeng, et~al.]{yang2023reco}
Zhengyuan Yang, Jianfeng Wang, Zhe Gan, Linjie Li, Kevin Lin, Chenfei Wu, Nan Duan, Zicheng Liu, Ce Liu, Michael Zeng, et~al.
\newblock Reco: Region-controlled text-to-image generation.
\newblock In \emph{CVPR}, pages 14246--14255, 2023{\natexlab{b}}.

\bibitem[Zhang and Agrawala(2024)]{zhang2024transparent}
Lvmin Zhang and Maneesh Agrawala.
\newblock Transparent image layer diffusion using latent transparency.
\newblock \emph{arXiv preprint arXiv:2402.17113}, 2024.

\bibitem[Zhang et~al.(2023{\natexlab{a}})Zhang, Rao, and Agrawala]{zhang2023adding}
Lvmin Zhang, Anyi Rao, and Maneesh Agrawala.
\newblock Adding conditional control to text-to-image diffusion models.
\newblock In \emph{ICCV}, pages 3836--3847, 2023{\natexlab{a}}.

\bibitem[Zhang et~al.(2023{\natexlab{b}})Zhang, Zhao, Lu, and Chien]{zhang2023text2layer}
Xinyang Zhang, Wentian Zhao, Xin Lu, and Jeff Chien.
\newblock Text2layer: Layered image generation using latent diffusion model.
\newblock \emph{arXiv preprint arXiv:2307.09781}, 2023{\natexlab{b}}.

\bibitem[Zheng et~al.(2023)Zheng, Zhou, Li, Qi, Shan, and Li]{zheng2023layoutdiffusion}
Guangcong Zheng, Xianpan Zhou, Xuewei Li, Zhongang Qi, Ying Shan, and Xi Li.
\newblock Layoutdiffusion: Controllable diffusion model for layout-to-image generation.
\newblock In \emph{CVPR}, pages 22490--22499, 2023.

\bibitem[Zheng et~al.(2024)Zheng, Gao, Fan, Liu, Laaksonen, Ouyang, and Sebe]{zheng2024bilateral}
Peng Zheng, Dehong Gao, Deng-Ping Fan, Li Liu, Jorma Laaksonen, Wanli Ouyang, and Nicu Sebe.
\newblock Bilateral reference for high-resolution dichotomous image segmentation.
\newblock \emph{arXiv preprint arXiv:2401.03407}, 2024.

\bibitem[Zou et~al.(2025)Zou, Feng, Wang, Huang, Huang, Haihang, Zou, and Li]{zou2025zero}
Kaifeng Zou, Xiaoyi Feng, Peng Wang, Tao Huang, Zizhou Huang, Zhang Haihang, Yuntao Zou, and Dagang Li.
\newblock Zero-shot subject-centric generation for creative application using entropy fusion.
\newblock \emph{arXiv preprint arXiv:2503.10697}, 2025.

\end{thebibliography}
}

\clearpage
\setcounter{page}{1}
\maketitlesupplementary

This supplementary material provides additional analysis and results that complement the main submission. Specifically, we delve deeper into the mechanisms and benefits of our proposed Training-Free Chroma Key Content Generation Diffusion Model (TKG-DM). We explore key aspects such as the positive-to-negative ratio in the channel mean shift, the denoising process, the impact of classifier-free guidance, and the effects of negative prompts. All content here is based on the methodology outlined in the main manuscript.

\section{Additional Analysis}
\label{sec:sup_analysis}
To further illustrate the capabilities of TKG-DM in generating chroma key content, we present comprehensive visualizations and analyses. These include examinations of the positive-to-negative ratio, detailed observations of the denoising process, an exploration of classifier-free guidance effects, and an assessment of negative prompts.

\subsection{Positive-to-Negative Ratio in Channel Mean Shift}
\label{sec:cms}
\begin{figure*}[t]
    \centering
    \includegraphics[width=0.85\textwidth]{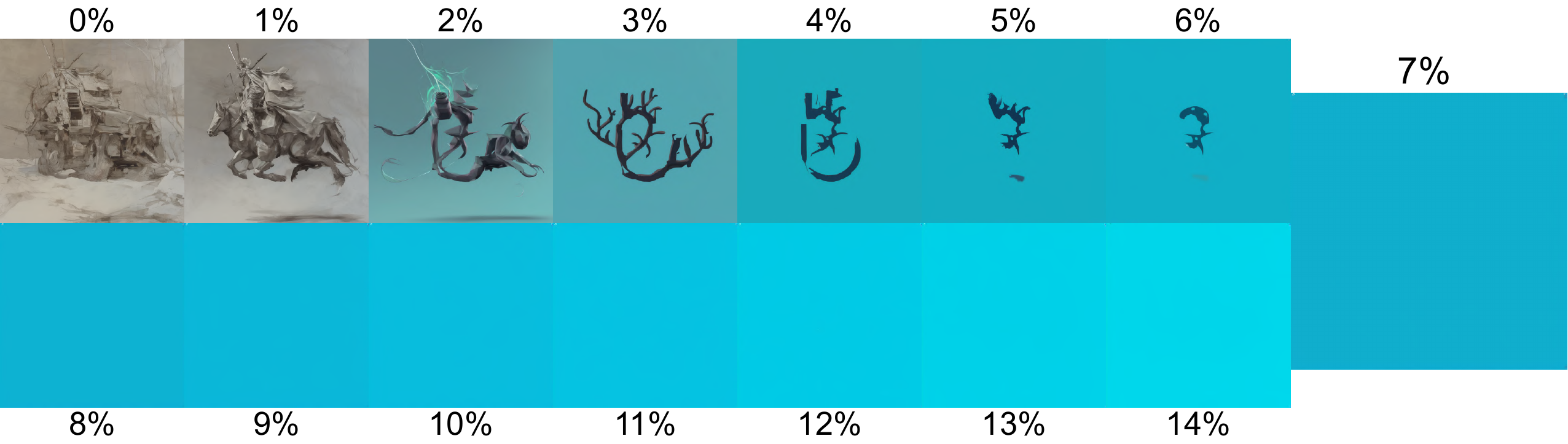} 
    \caption{Relationship between the positive-to-negative ratio and single-channel mean shift (channel = 2) without a text prompt. As the ratio increases, the background shifts towards a monochromatic blue.}
    \label{fig:ratio_wo_b}
\end{figure*}

\begin{figure*}[t]
    \centering
    \includegraphics[width=0.85\textwidth]{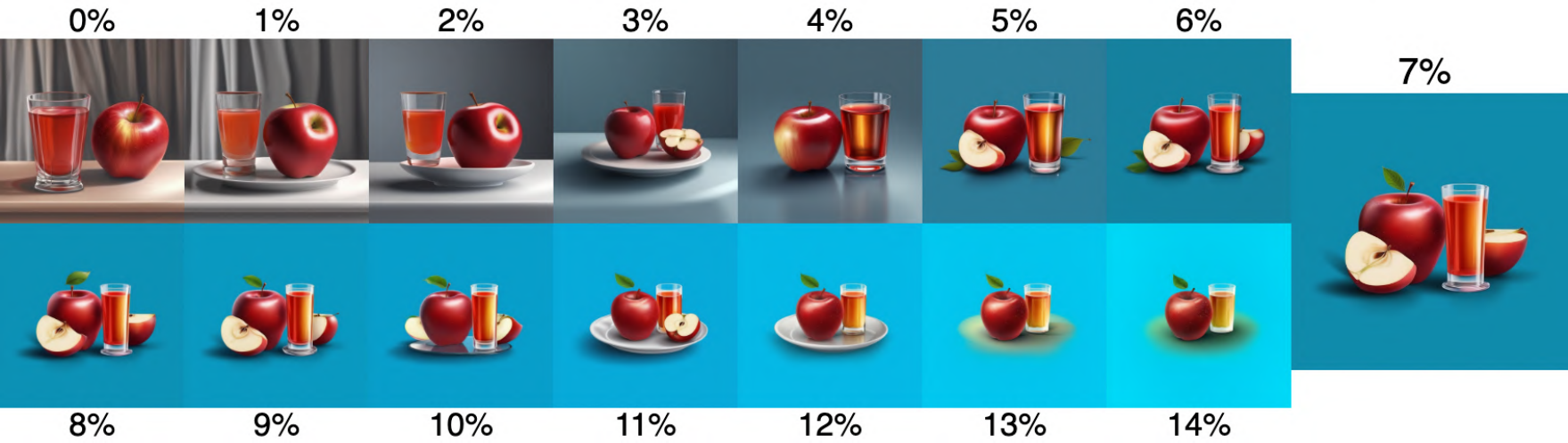} 
    \caption{Relationship between the positive-to-negative ratio and single-channel mean shift (channel = 2) with the text prompt ``red apple and glass of juice''. Higher ratios lead to a monochromatic blue background but may degrade foreground alignment with the prompt.}
    \label{fig:ratio_w_b}
\end{figure*}

\begin{figure*}[t]
    \centering
    \includegraphics[width=0.85\textwidth]{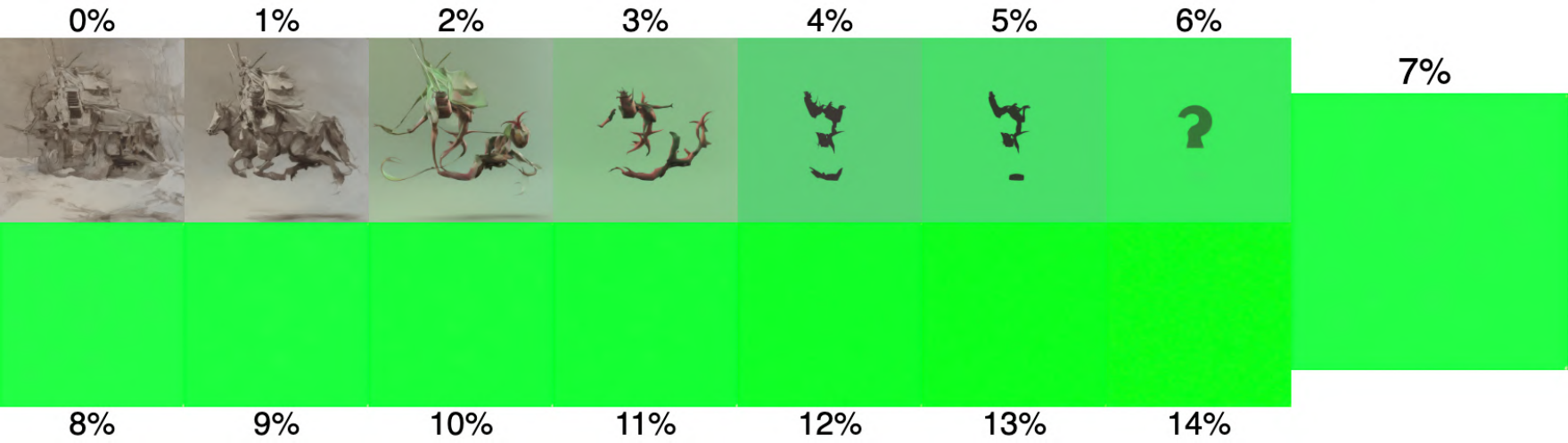} 
    \caption{Relationship between the positive-to-negative ratio and multi-channel mean shift (channels = 2 and 3) without a text prompt. Increasing the ratio produces a consistent monochromatic green background.}
    \label{fig:ratio_wo}
\end{figure*}

\begin{figure*}[t]
    \centering
    \includegraphics[width=0.85\textwidth]{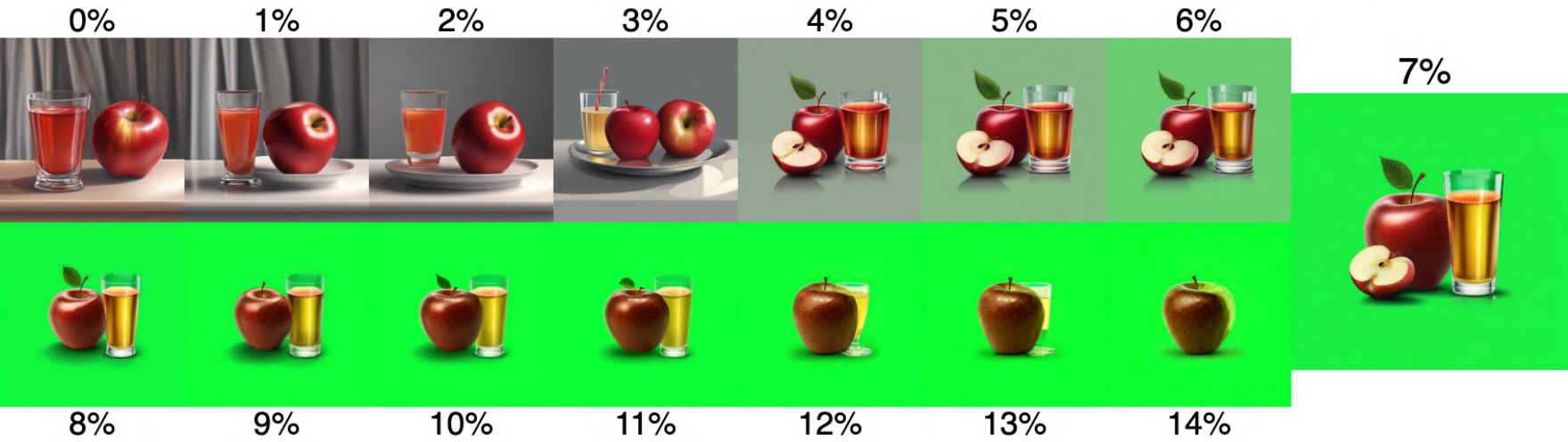} 
    \caption{Relationship between the positive-to-negative ratio and multi-channel mean shift (channels = 2 and 3) with the text prompt ``red apple and glass of juice''. Optimal ratios balance background uniformity and foreground alignment.}
    \label{fig:ratio_w}
\end{figure*}

To generate images with a specified background color, we introduce the channel mean shift technique as detailed in Section~\ref{sec:channe_mean_shift}. This method leverages the relationship between the positive-to-negative ratio of each channel in the initial noise and the resulting color in the generated image. By adjusting this ratio, we aim to optimize the initial noise to produce the desired initial color noise.
Fig.~\ref{fig:ratio_wo_b},~\ref{fig:ratio_w_b} and Fig.~\ref{fig:ratio_wo},~\ref{fig:ratio_w}  show results for both single-channel (channel = 2) and multi-channel (channels = 2 and 3) mean shifts for blue and green backgrounds, respectively. These figures demonstrate that as the positive-to-negative ratio increases, the method shifts toward generating a monochromatic background.

For cases without a prompt, both single-channel and multi-channel mean shifts exhibit a similar trend, where the foreground image disappears when the positive-to-negative ratio reaches approximately 7\%. In the single-channel scenario, increasing the mean shift results in progressively brighter tones. In contrast, the multi-channel case produces consistent monochromatic images beyond this threshold.

When a text prompt is provided, the foreground object is generated with strong semantic alignment to the prompt on the desired color background. However, if the positive-to-negative ratio is too low, the foreground aligns well with the text, but the background retains unwanted object features. Conversely, if the ratio is too high, the background becomes a solid color, but the semantic alignment of the foreground with the text deteriorates.

These results highlight the trade-off between foreground text alignment and background color uniformity. Our technique optimizes the initial noise distribution by carefully adjusting the positive-to-negative ratio, effectively balancing these factors. This enables the generation of consistent, high-quality chroma key backgrounds across various noise distributions, both single-channel and multi-channel.

\subsection{Denoising Process}
\label{sec:sup_denoise}
\textbf{Denoising Step.}
To control the size, position, and number of foreground objects, we introduce the init noise selection strategy in Section~\ref{sec:init_selector}. Fig.~\ref{fig:supp_step_normal},~\ref{fig:supp_step_gb} and ~\ref{fig:supp_step_ours} show visualizations of each denoising step for a standard SDXL, SDXL with a Green Background Prompt (GBP), and TKG-DM, respectively. As demonstrated in Fig.~\ref{fig:supp_step_gb} and ~\ref{fig:supp_step_ours}, using normal initial noise struggles to produce a stable single foreground object or generate a monochromatic background. In contrast, our noise setup enables Stable Diffusion to generate monochromatic background images across various denoising steps.
Specifically, with a small denoising step value ($\text{denoising step} = 5$), our approach produces coherent semantic structures. These results indicate that our method enables high-precision chroma key image generation even with fewer denoising steps compared to normal initial noise.
Further analysis at a denoising step value of 1 shows that, with the introduction of initial color noise, the model produces images with a prominent green element in the background, unlike when using normal initial noise. This demonstrates that our method effectively generates background color elements, even with minimal denoising steps.

\textbf{Intermediate Step.} Next, we examine the intermediate results during the denoising process when the total number of denoising steps is set to 50. For a clear comparison, we conducted this analysis using three configurations: normal SDXL, SDXL with a Green Background Prompt (GBP), and TKG-DM, as shown in Fig. \ref{fig:supp_step_normal}, Fig. \ref{fig:supp_step_gb}, and Fig. \ref{fig:supp_step_ours}, respectively. At an early denoising step (e.g., step = 1), our method exhibits a stronger green component than the normal initial noise, indicating that our initial noise adjustments effectively guide the generation process toward a green chroma key background from the outset.

\subsection{Classifier-Free Guidance}
\label{sec:sup_cfg}
\begin{figure*}[t]
    \centering
    \includegraphics[width=0.9\textwidth]{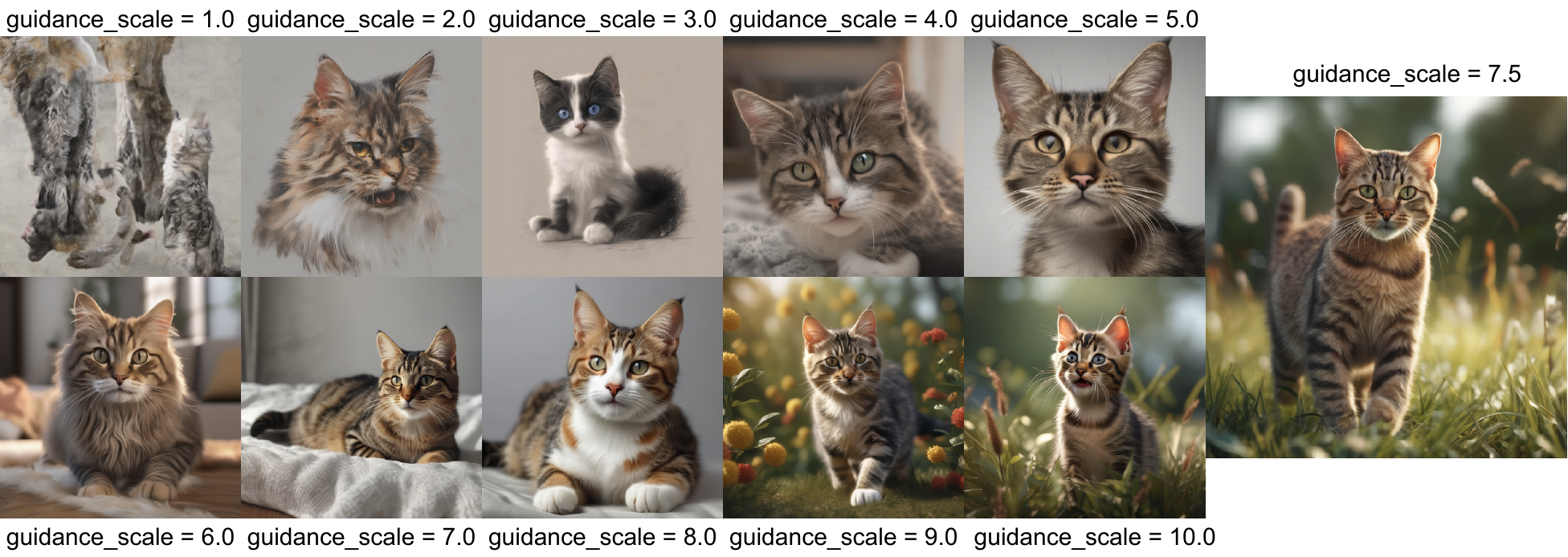} 
    \caption{Effect of varying guidance scales in SDXL. The guidance scale ranges from 1.0 to 10.0. The input prompt is ``The cat''. Higher guidance scales improve text alignment but may introduce background artifacts.}
    \label{fig:supp_cfg_normal}
\end{figure*}

\begin{figure*}[t]
    \centering
    \includegraphics[width=0.9\textwidth]{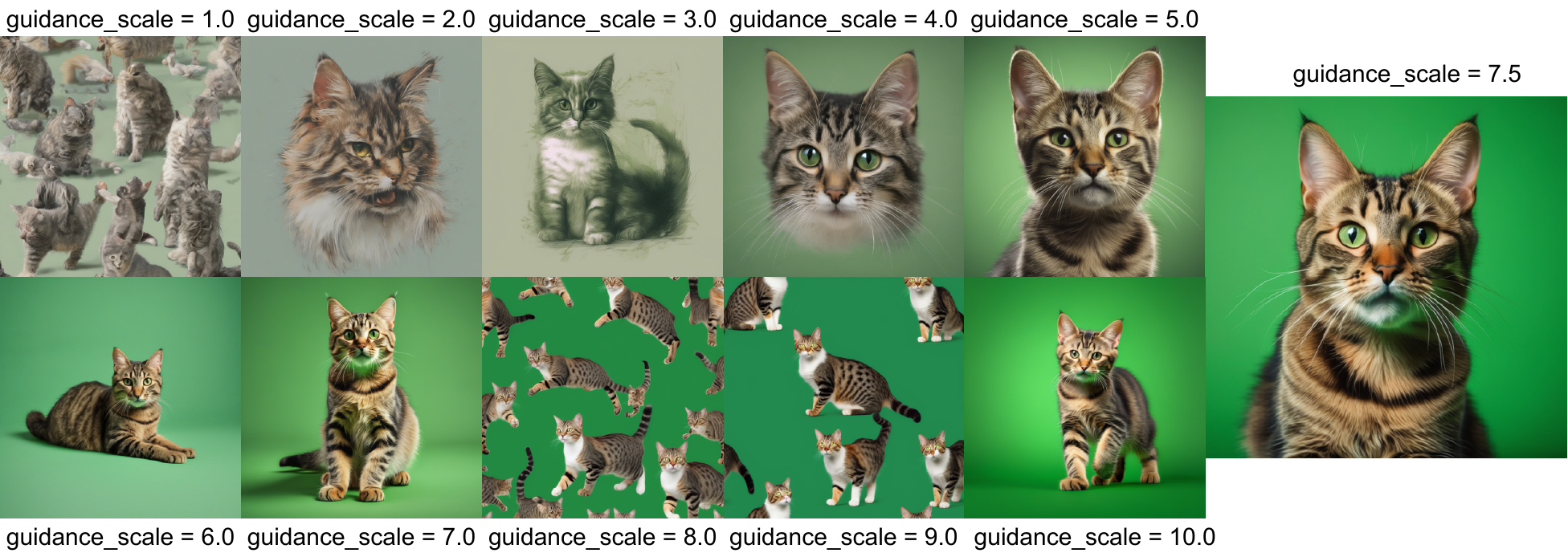} 
    \caption{Effect of varying guidance scales in SDXL with Green Background Prompt. The input prompt is ``The cat''. Increasing the guidance scale enhances the green background but may affect foreground details.}
    \label{fig:supp_cfg_gb}
\end{figure*}

\begin{figure*}[t]
    \centering
    \includegraphics[width=0.9\textwidth]{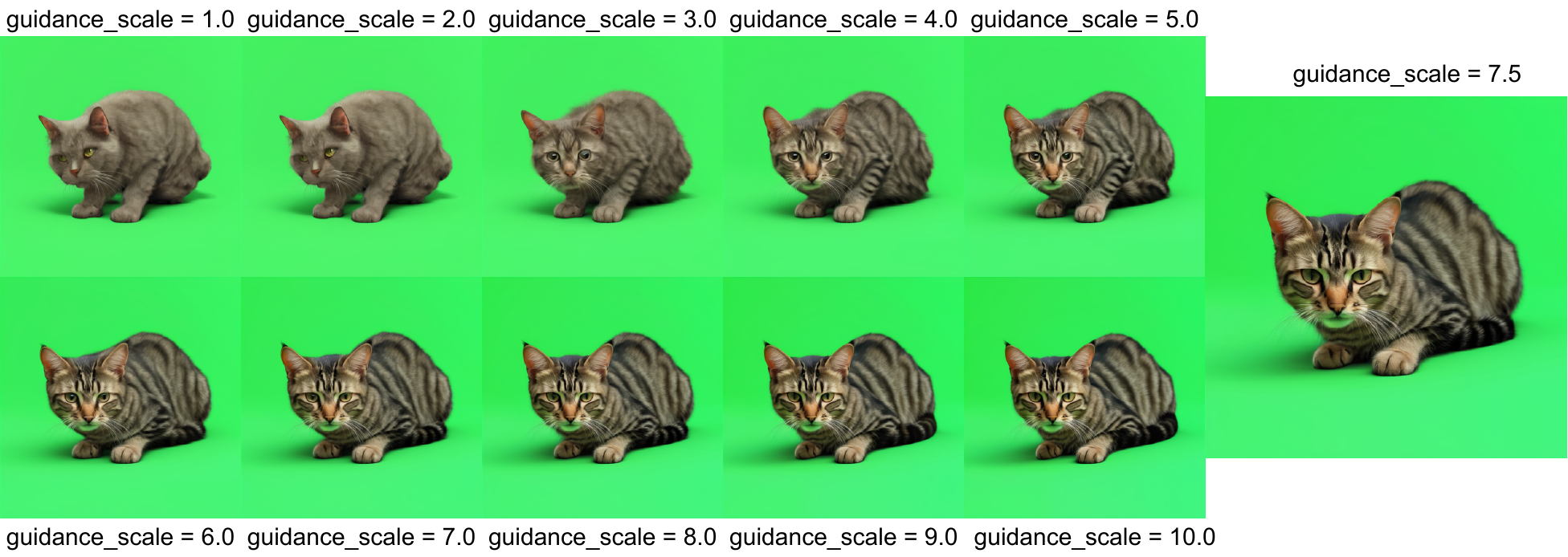} 
    \caption{Effect of varying guidance scales in TKG-DM. The input prompt is ``The cat''. Our method maintains consistent background and foreground quality regardless of the guidance scale.}
    \label{fig:supp_cfg_ours}
\end{figure*}

Then, we analyze the relationship between TKG-DM and Classifier-Free Guidance (CFG) by adjusting the guidance scale values. Fig.~\ref{fig:supp_cfg_normal}, Fig.~\ref{fig:supp_cfg_gb} and Fig.~\ref{fig:supp_cfg_ours} show results for normal SDXL, SDXL with the Green Background Prompt (GBP), and TKG-DM, respectively.
For SDXL with GBP, increasing the guidance scale enhances the prominence of the green background. In contrast, TKG-DM maintains a consistent monochromatic background, unaffected by changes in guidance scale, due to the background control provided by init color noise. This finding demonstrates that in TKG-DM, the background color is independent of the text prompt and unaffected by the guidance scale (CFG). In other words, the background color information relies solely on the init color noise, enabling the generation of monochromatic backgrounds without any dependency on the text prompt.

\subsection{Negative Prompt} 
\label{sec:sup_np}
\begin{figure*}[t]
    \centering
    \includegraphics[width=0.8\textwidth]{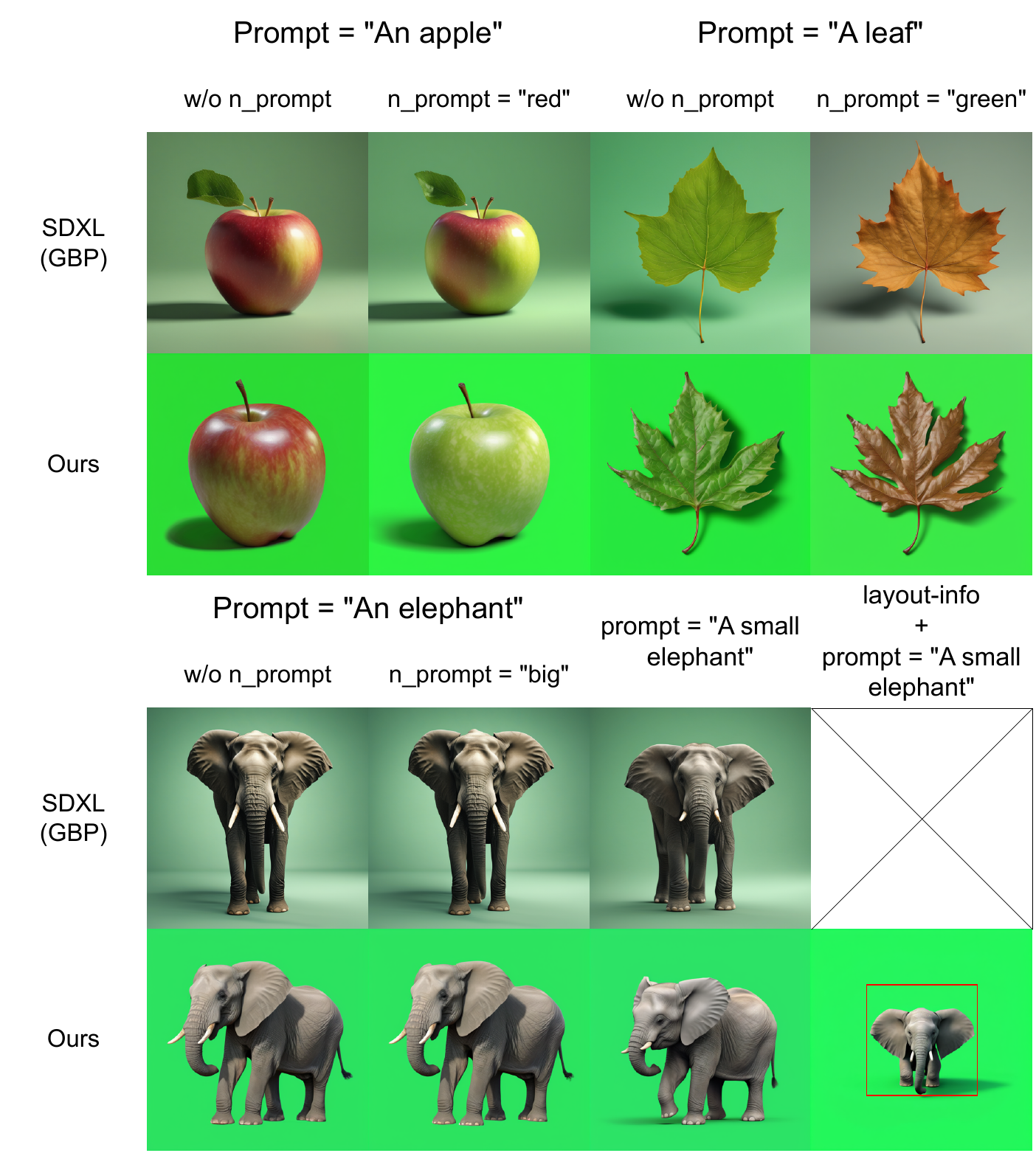} 
    \caption{Comparison of results using negative prompts to modify colors and control size in generated images. For color adjustments, the negative prompt removes specific tones, such as green, in both SDXL with Green Background Prompt (GBP) and TKG-DM. In TKG-DM, init color noise enables selective color removal without affecting the background, demonstrating improved control. However, for size control, where negative prompts are generally ineffective, TKG-DM achieves size adjustments through init noise selection, providing enhanced flexibility in generating objects like an elephant at various scales.}
    \label{fig:supp_np}
\end{figure*}

To further examine TKG-DM’s control capabilities, we applied a negative prompt to remove green tones from the foreground. As shown in Fig. \ref{fig:supp_np}, we compared results for SDXL with Green Background Prompt (GBP) and TKG-DM. In SDXL with GBP, the background prompt’s green setting interferes with the effectiveness of the negative prompt, limiting control over the foreground’s green tones. However, in TKG-DM, where init color noise controls the background, the negative prompt selectively influences only the foreground. This demonstrates TKG-DM’s superior ability to manage negative prompts for precise color control in the foreground without affecting the background.

Additionally, when generating an elephant, traditional models struggle to adjust size using either prompts or negative prompts. In contrast, as discussed in Section~\ref{sec:application}, TKG-DM allows users to adjust object size more directly, providing a significant advantage in generating objects at varying scales as needed.

\section{Additional Results}
\label{sec:sup_ad_result}
\subsection{Additional Results: TKG-DM with Green Chroma Key Backgrounds}
\label{sec:sup_result}
\begin{figure*}[t]
    \centering
    \includegraphics[width=0.8\textwidth]{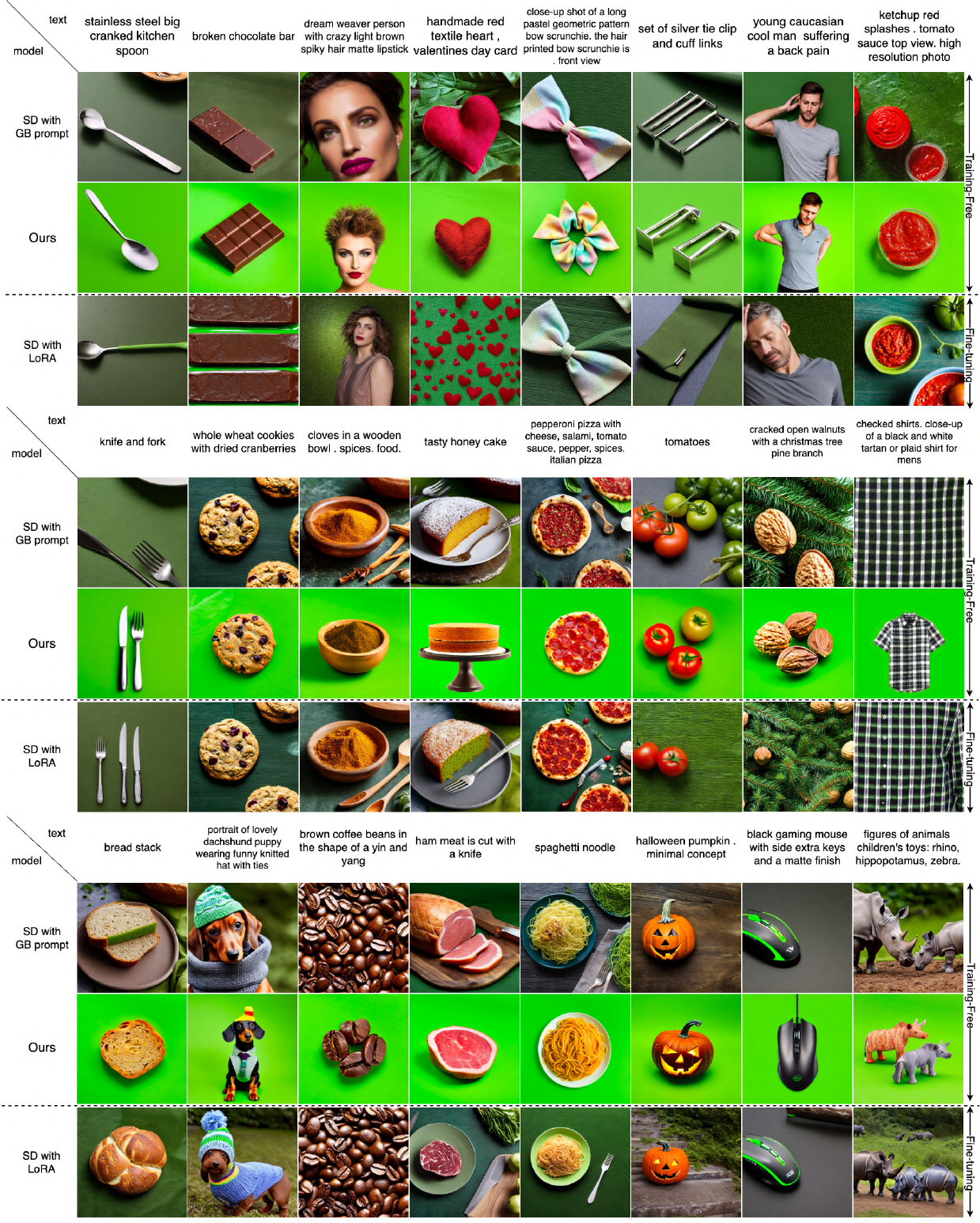} 
    \caption{Additional Result in Green Background with SD1.5}
    \label{fig:supp_ad_result_15}
\end{figure*}

\begin{figure*}[t]
    \centering
    \includegraphics[width=0.78\textwidth]{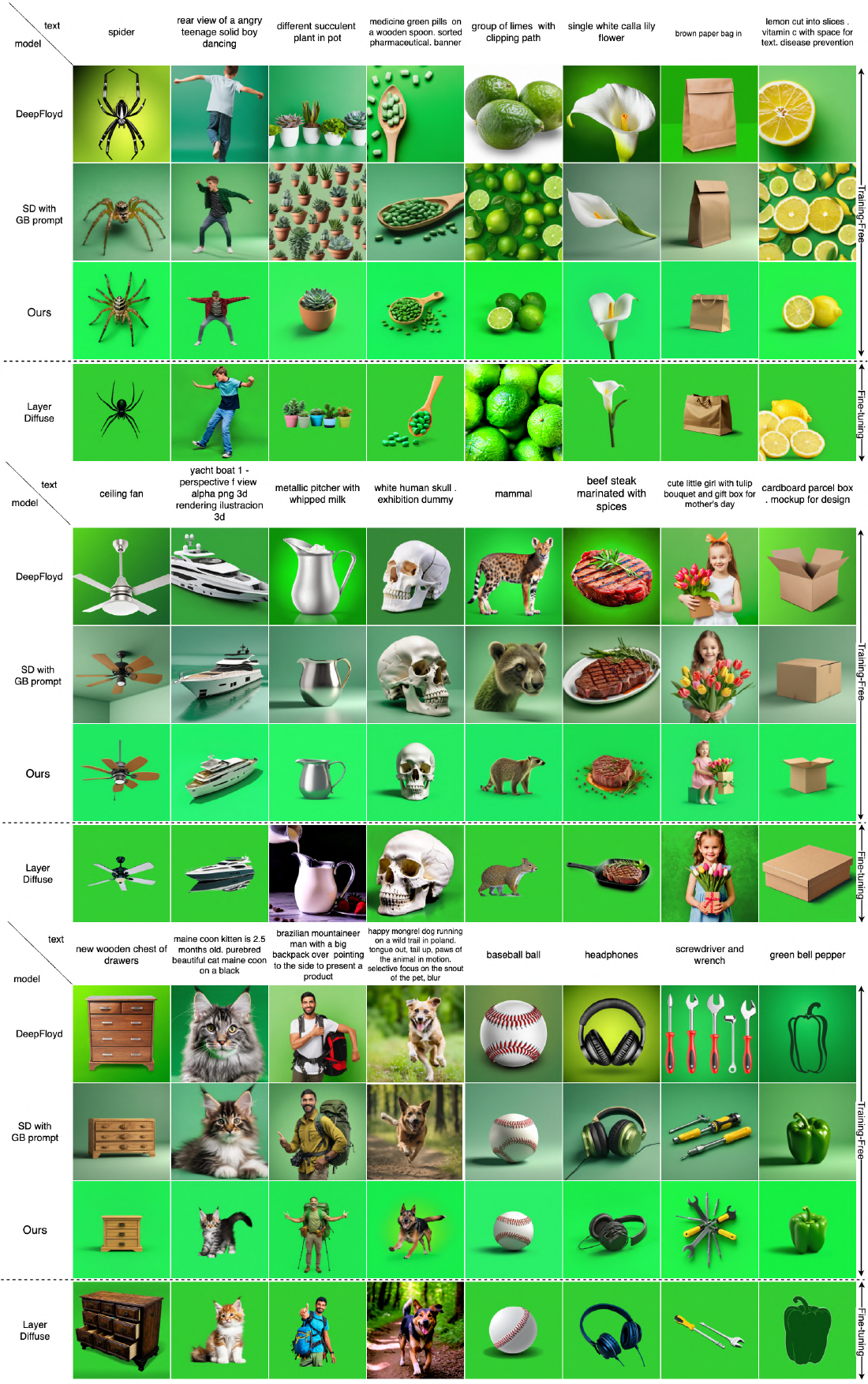} 
    \caption{Additional Result in Green Background with SDXL}
    \label{fig:supp_ad_result_xl}
\end{figure*}

Fig.~\ref{fig:supp_ad_result_15} and Fig.~\ref{fig:supp_ad_result_xl} provide additional examples based on SD1.5 and SDXL, demonstrating TKG-DM’s effectiveness in generating high-quality, chroma-keyed foreground images against a green background.

\subsection{Additional Results: TKG-DM with Various Chroma Key Backgrounds}
\label{sec:sup_various}
TKG-DM allows the user to control the background via the selection channel in channel mean shift (Section~\ref{sec:analysis}). 
Fig.~\ref{fig:result_color} provides additional examples of chroma key backgrounds in various colors. This demonstrates our approach’s effectiveness in generating diverse chroma key backgrounds.

\subsection{Additional Results: TKG-DM with Flow-based Model} \label{sec:sup_flow} 
\begin{figure*}[t] \centering 
\includegraphics[width=0.65\textwidth]{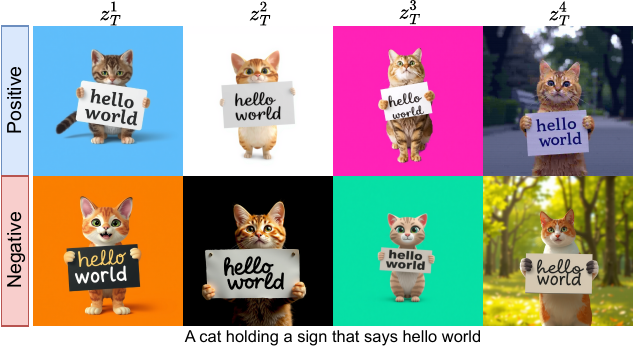} \caption{\textbf{TKG-DM applied to the FLUX flow-based model.} By adjusting the mean shift in specific channels, TKG-DM generates green and blue backgrounds without fine-tuning the model itself. The differences in each channel’s color representation from SD1.5 and SDXL highlight TKG-DM’s architecture-agnostic design.} \label{fig:flux_demo} \end{figure*}

To demonstrate the generality of our approach beyond standard Stable Diffusion (SD) variants, we also applied TKG-DM to a flow-based matching model, FLUX~\cite{lipman2022flow}. As in our SD-based experiments, we \emph{freeze} the FLUX model parameters and leverage the same Channel Mean Shift technique, highlighting its training-free and wide applicability. Fig.~\ref{fig:flux_demo} presents example results produced by FLUX with a specified background color using our channel mean shift strategy.

Interestingly, FLUX encodes color information into its latent channels in a way that diverges from SD1.5 or SDXL. Channel 1 often correlates with blue and orange hues, Channel 2 predominantly captures black and white tones, Channel 3 represents pink and green relationships, and Channel 4 governs brightness factors. Although these roles contrast with the channel distributions in SD-based models, our framework remains effective because it only requires identifying which channels to shift for the desired color effect. This flexibility further illustrates that TKG-DM is not tied to any specific network architecture or parameter distribution, and can be seamlessly transferred to other generative pipelines—including flow-based, diffusion-based, or new emerging frameworks—without additional model fine-tuning. Consequently, TKG-DM holds promise for a wide range of downstream tasks where reliable chroma-key content generation is needed, regardless of the underlying generative model’s channel semantics.

\section{More Results of TKG-DM with application track}
\label{sec:sup_application}
Beyond text-to-image tasks, TKG-DM enables chroma key content generation across various applications. The main paper highlights applications involving ControlNet, layout-aware text-to-image, consistency models, and text-to-video models. Additionally, we demonstrate applications of TKG-DM with Multi-Diffusion and text-to-3D generation.

\subsection{Application to Multi-Diffusion}
Multi-Diffusion~\cite{bartal2023multidiffusionfusingdiffusionpaths} enables the generation of different image regions using distinct diffusion processes. By applying TKG-DM to the central portion of the image while keeping other areas under normal noise conditions, we can control which parts of the image have chroma key backgrounds. This flexibility allows precise control over both background and foreground regions, supporting seamless integration into multi-object scene generation and interactive content creation.

\subsection{Application to Text-to-3D}
Using SV3D~\cite{voleti2024sv3d} with TKG-DM, we address text-to-3D generation. Although a complete text-to-3D pipeline is not yet implemented, TKG-DM’s background-free image generation facilitates efficient 3D object extraction without segmentation models, streamlining the creation of clean 3D assets from text prompts. Additionally, TKG-DM offers potential for conditional text-to-3D applications, similar to ControlNet, enabling enhanced control over 3D object generation based on specific textual inputs. This expands the possibility of creating customized 3D models that are aligned with user-defined criteria.

\subsection{Additional Results in Application}
\begin{figure*}[t]
    \centering
    \includegraphics[width=0.9\textwidth]{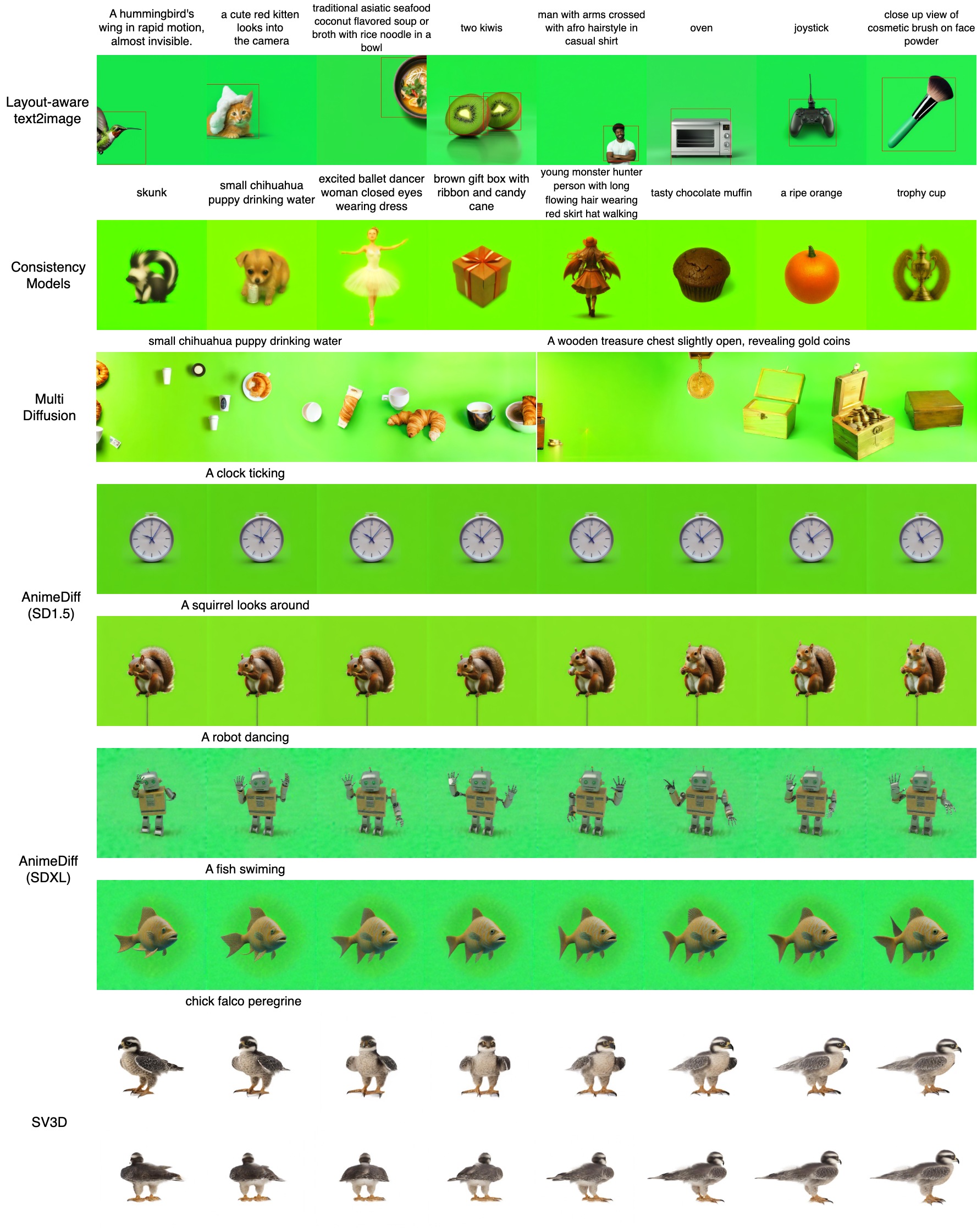} 
    \caption{Additional Result of Application track}
    \label{fig:supp_application}
\end{figure*}

Fig.~\ref{fig:supp_application} shows additional examples, including multi-diffusion and text-to-3D techniques applied without fine-tuning. We also present results where the chroma key processing was applied to foreground generation, allowing efficient and flexible background generation without additional training. These results highlight TKG-DM’s versatility across multiple generative tasks.


\begin{figure*}[t]
    \centering
    \includegraphics[width=0.9\textwidth]{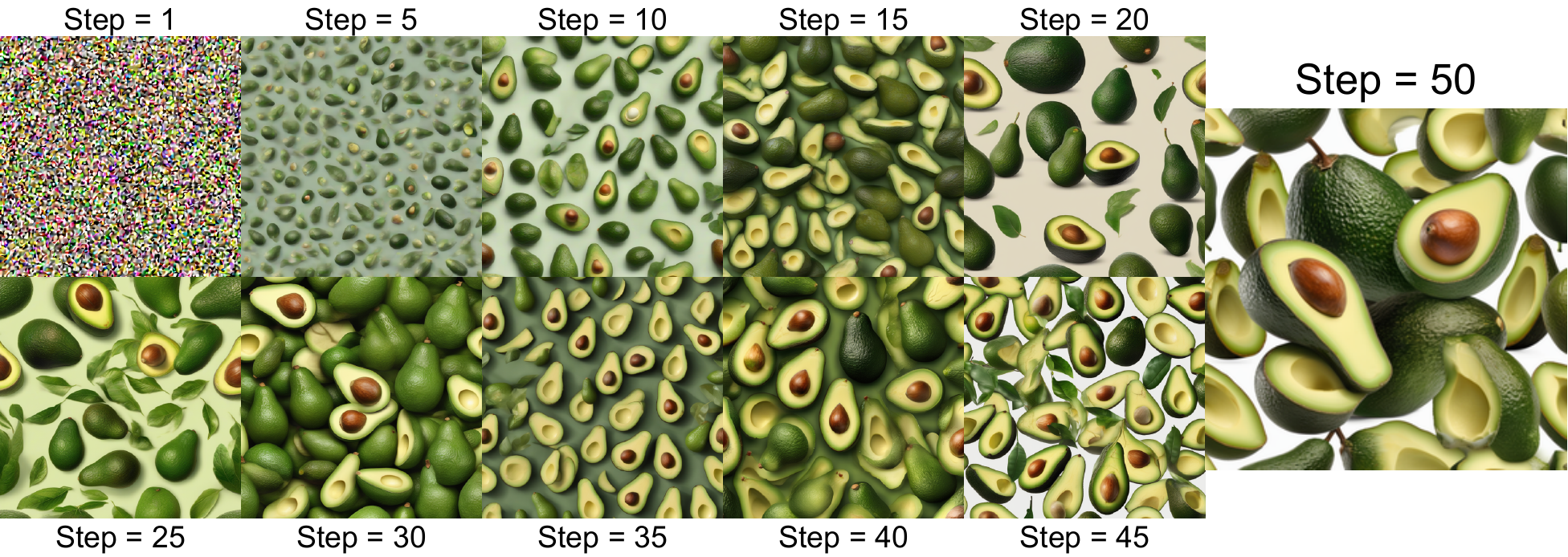} 
    \caption{Generated images in SDXL at various denoising steps (1 to 50). The images demonstrate the progression of generated content. The input prompt is ``An avocado''.}
    \label{fig:supp_step_normal}
\end{figure*}
\begin{figure*}[t]
    \centering
    \includegraphics[width=0.9\textwidth]{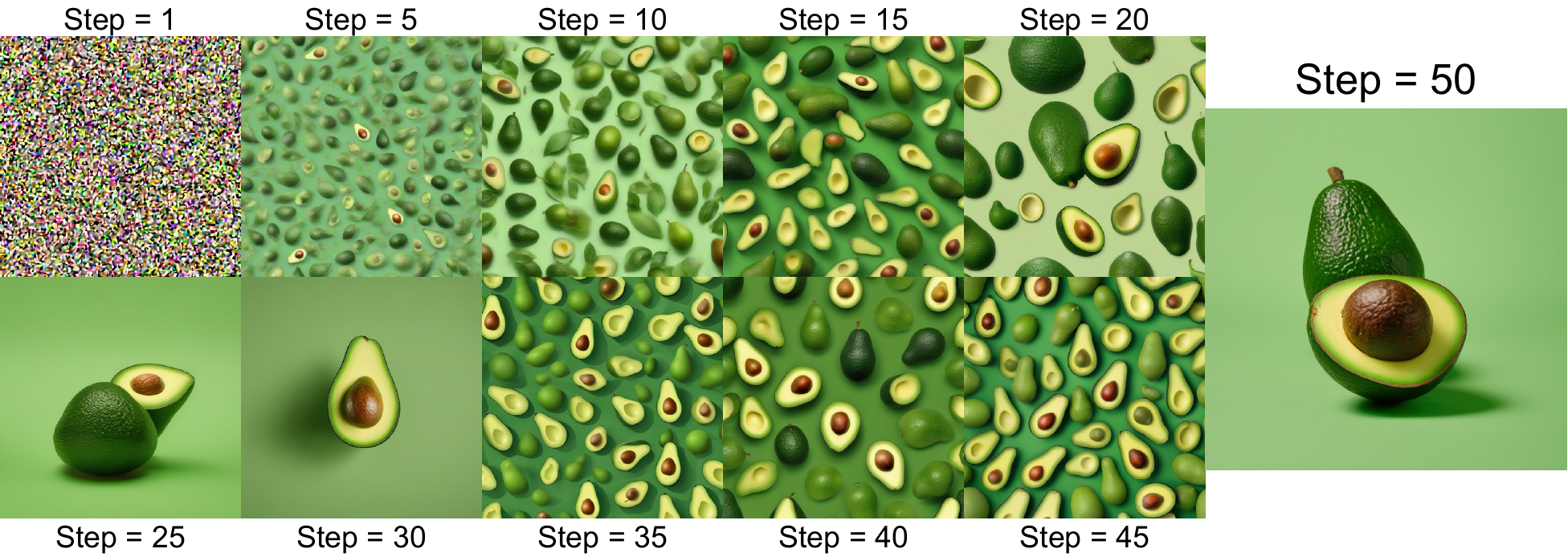} 
    \caption{Generated images in SDXL with Green Background Prompt at various denoising steps (1 to 50). The input prompt is ``An avocado''. The green background becomes more prominent with more denoising steps but may affect foreground quality.}
    \label{fig:supp_step_gb}
\end{figure*}

\begin{figure*}[t]
    \centering
    \includegraphics[width=0.9\textwidth]{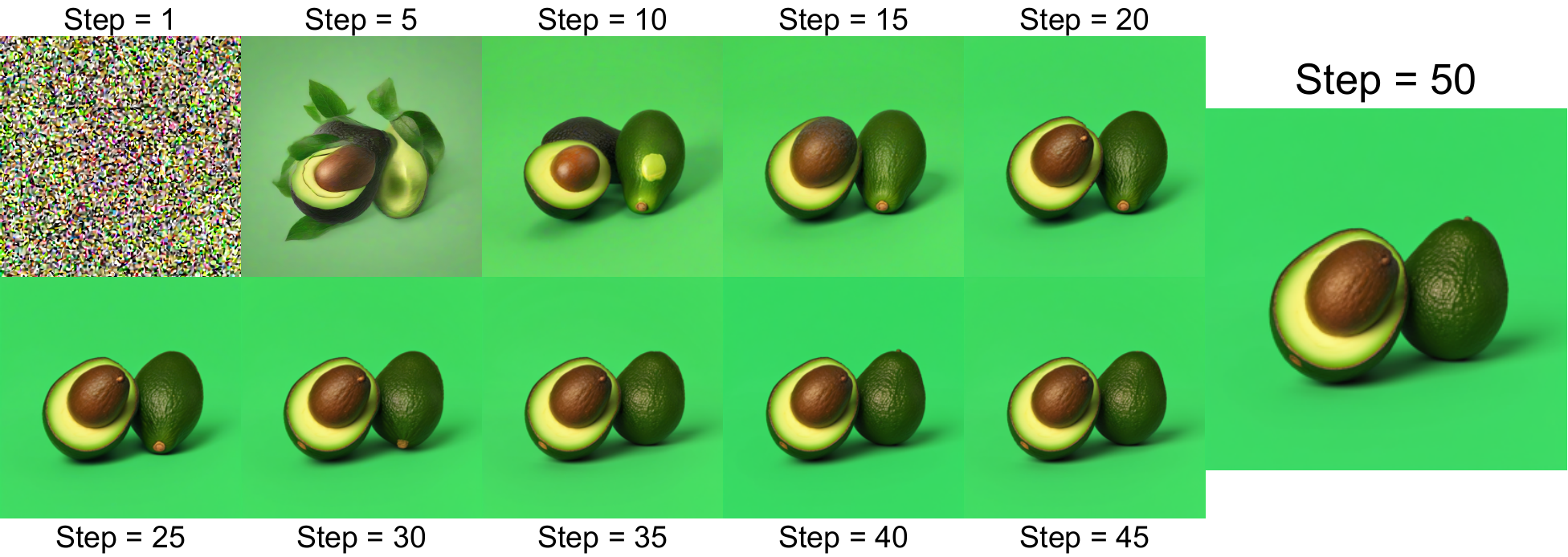} 
    \caption{Generated images using TKG-DM at various denoising steps (1 to 50). The input prompt is ``An avocado''. Our method maintains a consistent green background across all steps.}
    \label{fig:supp_step_ours}
\end{figure*}

\begin{figure*}[t]
    \centering
    \includegraphics[width=0.9\textwidth]{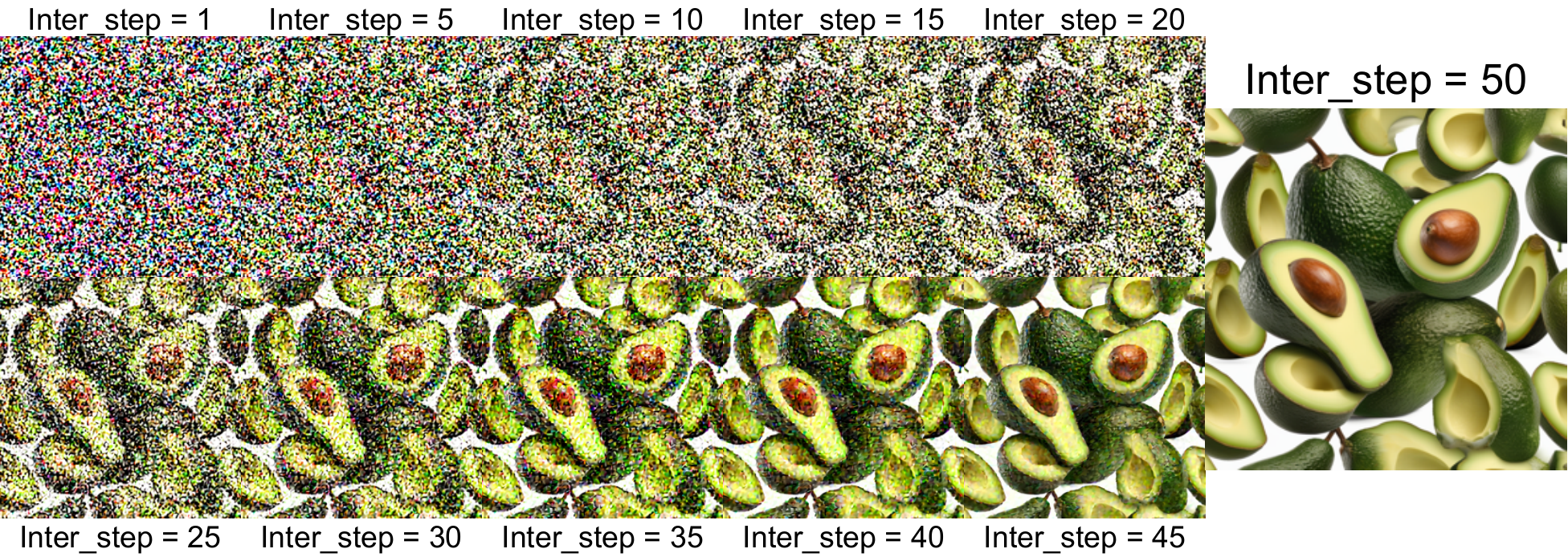} 
    \caption{Denoising process progression in SDXL at various steps. The images demonstrate the evolution of generated content from initial noise (Step = 1) to the final output (Step = 50). The input prompt = ``An avocado''.}
    \label{fig:supp_step_normal}
\end{figure*}
\begin{figure*}[t]
    \centering
    \includegraphics[width=0.9\textwidth]{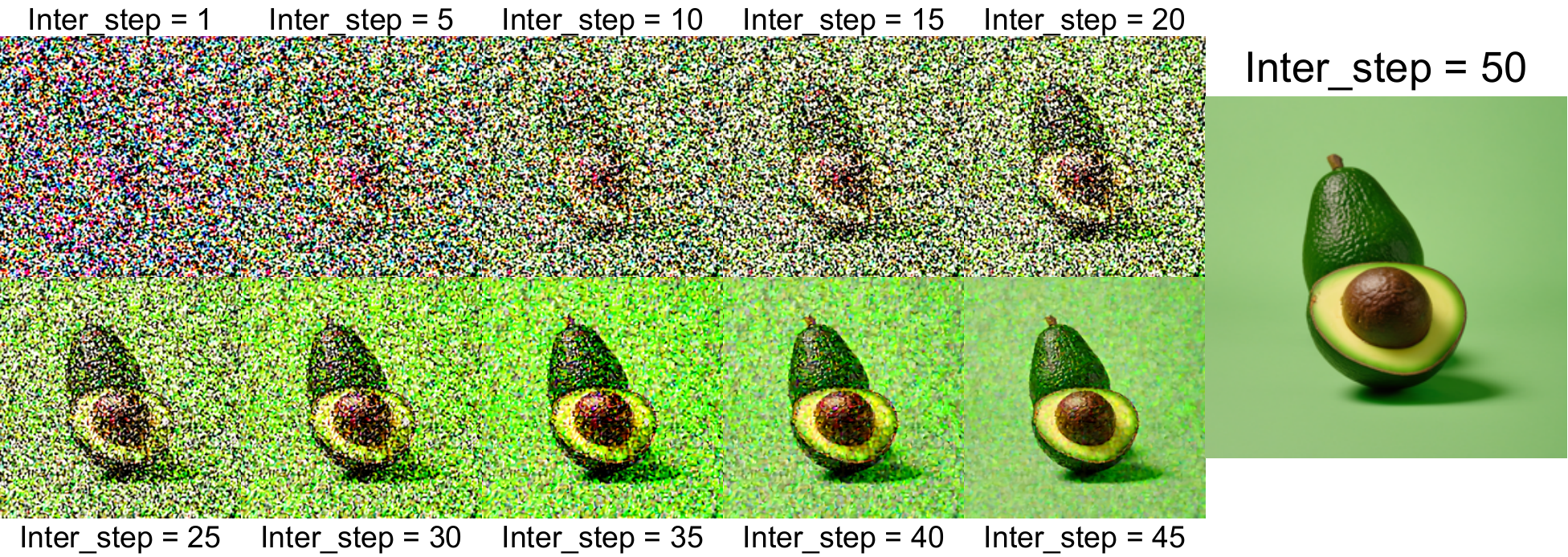} 
    \caption{Denoising process progression in SDXL with Green Background Prompt at various steps. The images demonstrate the evolution of generated content from initial noise (Step = 1) to the final output (Step = 50). The input prompt = ``An avocado''.}
    \label{fig:supp_step_gb}
\end{figure*}

\begin{figure*}[t]
    \centering
    \includegraphics[width=0.9\textwidth]{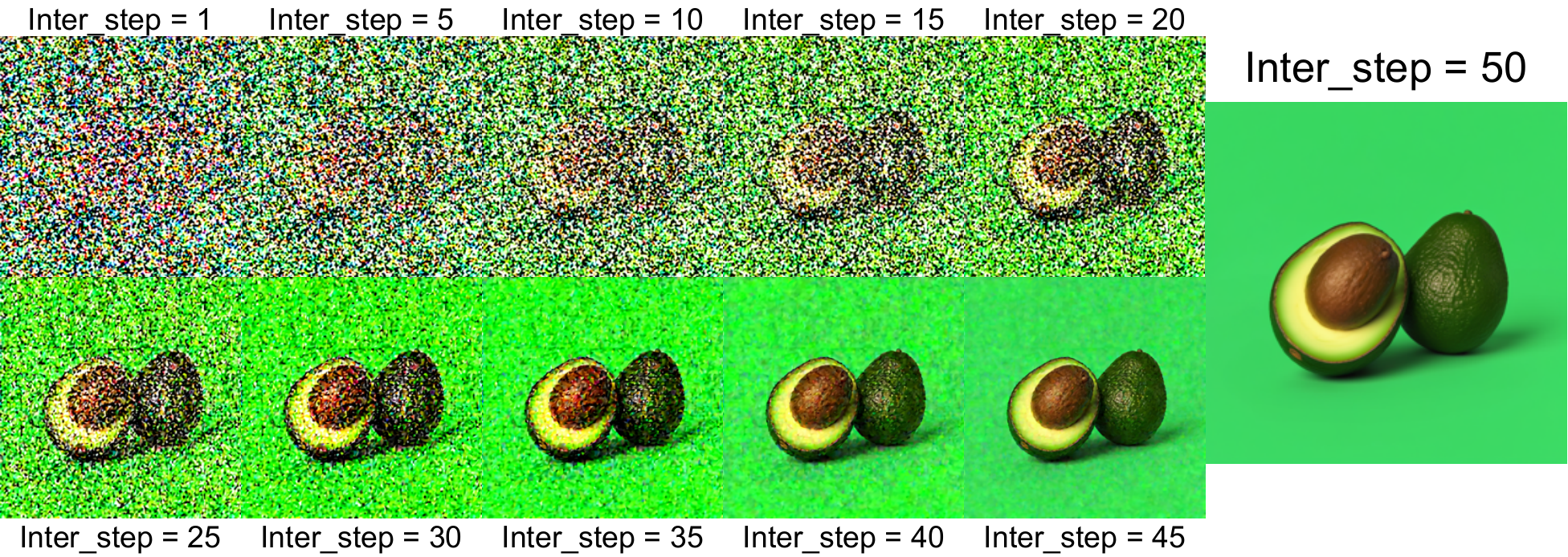} 
    \caption{Denoising process progression in our TKG-DM at various steps. The images demonstrate the evolution of generated content from initial noise (Step = 1) to the final output (Step = 50). The input prompt = ``An avocado''.}
    \label{fig:supp_step_ours}
\end{figure*}



\begin{figure*}[t]
    \centering
    \includegraphics[width=0.85\textwidth]{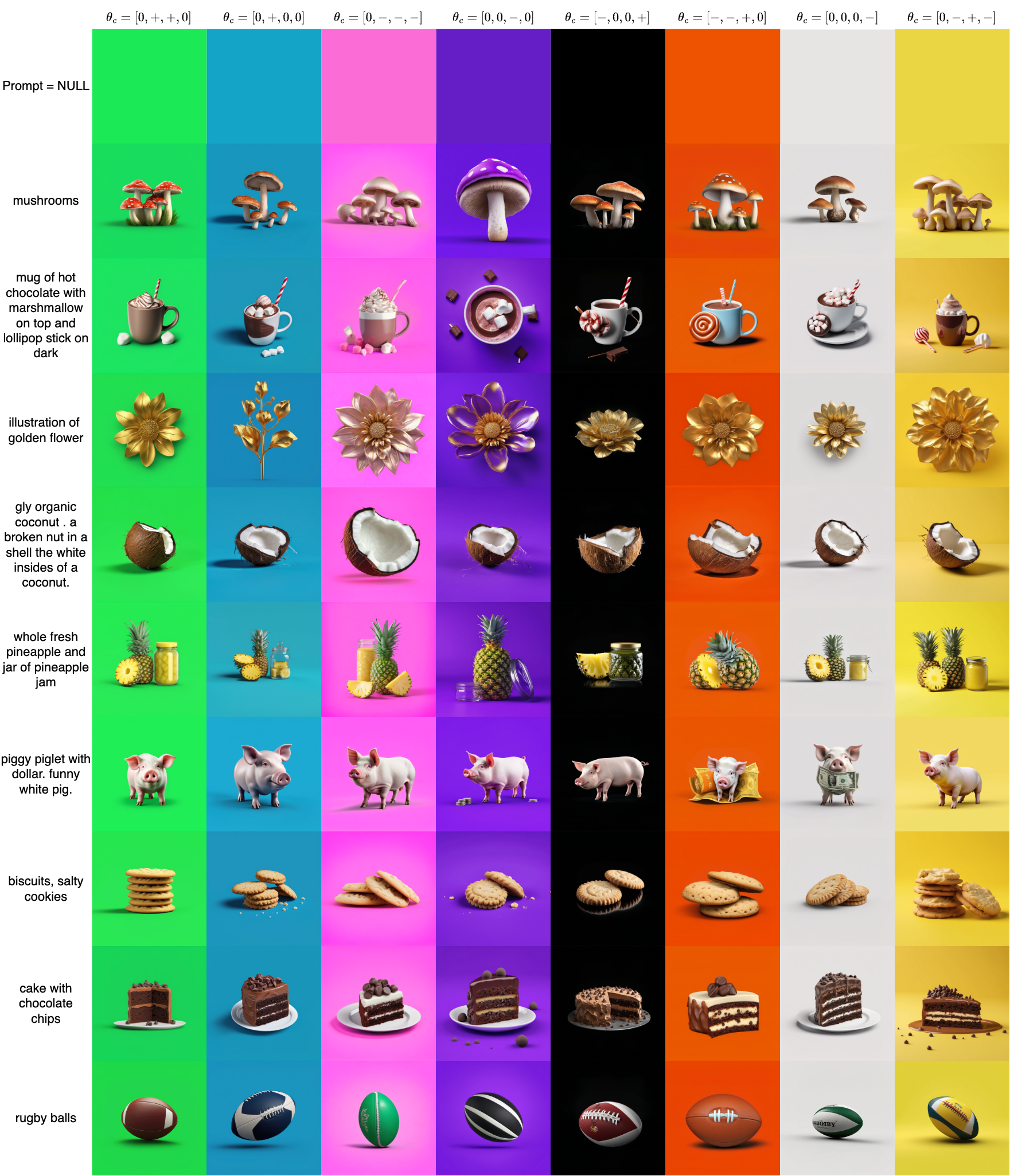} 
    \caption{Additional Result in various color Background with SDXL}
    \label{fig:supp_color}
\end{figure*}

\begin{figure*}[t]
    \centering
    \includegraphics[width=0.7\textwidth]{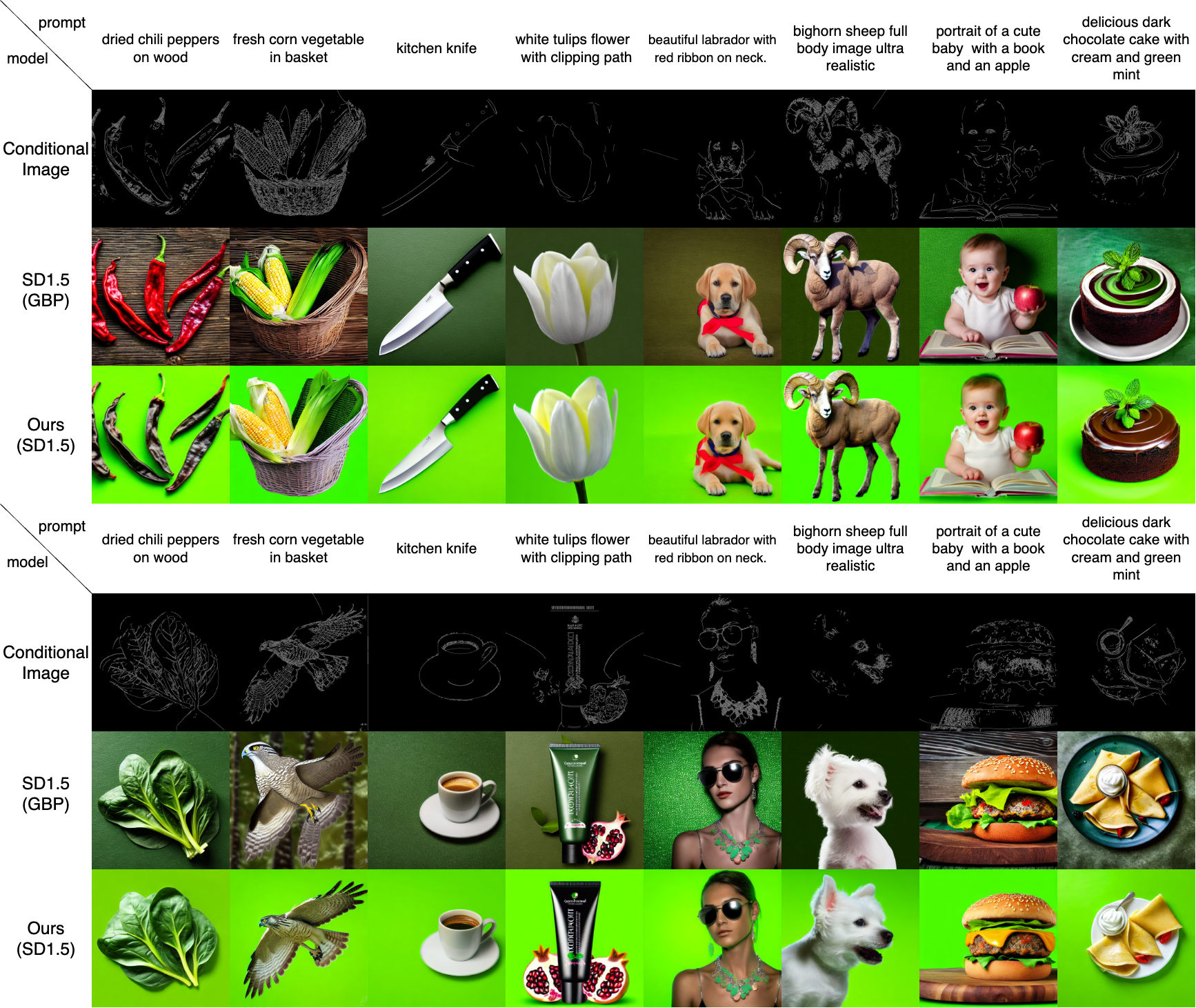} 
    \caption{Additional Result in ControlNet with SD1.5}
    \label{fig:supp_control15}
\end{figure*}

\begin{figure*}[t]
    \centering
    \includegraphics[width=0.7\textwidth]{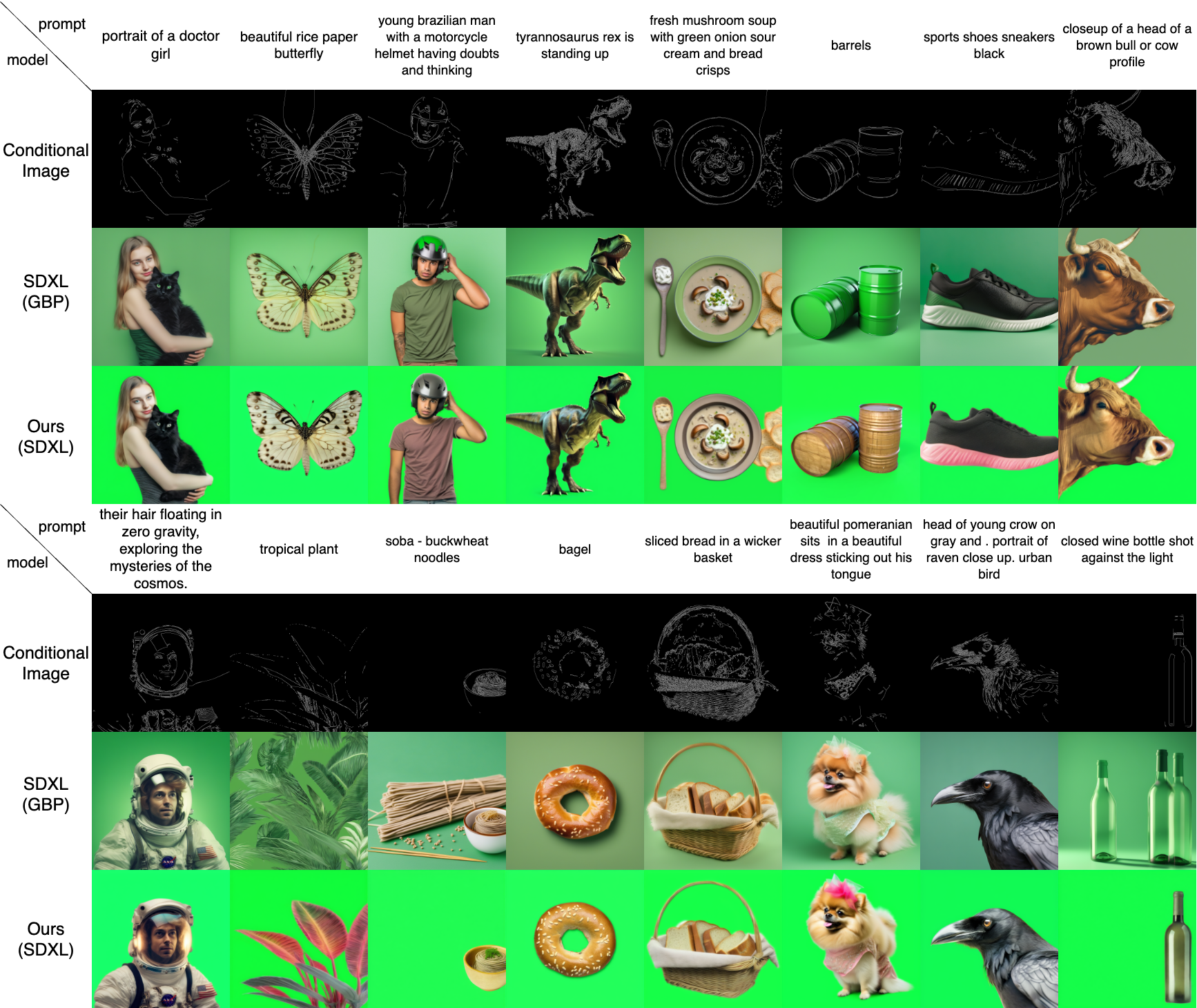} 
    \caption{Additional Result in ControlNet with SDXL}
    \label{fig:supp_controlxl}
\end{figure*}

\end{document}